\documentclass[lettersize,journal,x11names]{IEEEtran}
\usepackage{colortbl} 
\usepackage{array} 
\usepackage[x11names]{xcolor}
\usepackage{amsmath,amsfonts}
\usepackage{algorithmic}
\usepackage{array}
\usepackage[caption=false,font=normalsize,labelfont=sf,textfont=sf]{subfig}
\usepackage{textcomp}
\usepackage{stfloats}
\usepackage{url}
\usepackage{verbatim}
\usepackage{graphicx}
\usepackage{booktabs}
\usepackage{multirow}
\usepackage{mathtools}
\usepackage{amssymb} 

\def\BibTeX{{\rm B\kern-.05em{\sc i\kern-.025em b}\kern-.08em
  T\kern-.1667em\lower.7ex\hbox{E}\kern-.125emX}}
\usepackage{balance}
\begin{document}
\title{Modality-Collaborative Low-Rank Decomposers for Few-Shot Video Domain Adaptation}

\author{Yuyang~Wanyan, Xiaoshan~Yang, Weiming~Dong, and~Changsheng~Xu,~\IEEEmembership{Fellow,~IEEE}

\thanks{
This work was supported by the National Natural Science Foundation of China (Grants 62322212, U23A20387), the Beijing Natural Science Foundation (No. L221013), and the CAS Project for Young Scientists in Basic Research (YSBR-116). 

Yuyang Wanyan, Xiaoshan Yang, Weiming Dong and Changsheng Xu are with the State Key Laboratory of Multimodal Artificial Intelligence Systems, Institute of Automation, Chinese Academy of Sciences,
Beijing 100190, China, and the School of Artificial Intelligence, University
of Chinese Academy of Sciences, Beijing 100049, China. Xiaoshan Yang and Changsheng Xu are also with the PengCheng Laboratory, Shenzhen 518066, China (e-mail: wanyanyuyang2021@ia.ac.cn, xiaoshan.yang@nlpr.ia.ac.cn, weiming.dong@ia.ac.cn,
csxu@nlpr.ia.ac.cn).
}
}

\markboth{Journal of \LaTeX\ Class Files,~Vol.~18, No.~9, September~2020}%
{How to Use the IEEEtran \LaTeX \ Templates}

\maketitle

\begin{abstract}
In this paper, we study the challenging task of Few-Shot Video Domain Adaptation (FSVDA). 
The multimodal nature of videos introduces unique challenges, necessitating the simultaneous consideration of both domain alignment and modality collaboration in a few-shot scenario, which is ignored in previous literature. 
{\color{black}
We observe that, under the influence of domain shift, the generalization performance on the target domain of each individual modality, as well as that of fused multimodal features, is constrained. 
Because each modality is comprised of coupled features with multiple components that exhibit different domain shifts. 
This variability increases the complexity of domain adaptation, thereby reducing the effectiveness of multimodal feature integration. 
}
To address these challenges, we introduce a novel framework of Modality-Collaborative Low-Rank Decomposers (MC-LRD) to decompose modality-unique and modality-shared features with different domain shift levels from each modality that are more friendly for domain alignment. 
The MC-LRD comprises multiple decomposers for each modality and Multimodal Decomposition Routers (MDR).
{\color{black}
Each decomposer has progressively shared parameters across different modalities. 
The MDR is leveraged to selectively activate the decomposers to produce modality-unique and modality-shared features. }
To ensure efficient decomposition, we apply orthogonal decorrelation constraints separately to decomposers and sub-routers, enhancing their diversity. 
Furthermore, we propose a cross-domain activation consistency loss to guarantee that target and source samples of the same category exhibit consistent activation preferences of the decomposers, thereby facilitating domain alignment. 
Extensive experimental results on three public benchmarks demonstrate that our model achieves significant improvements over existing methods. 
\end{abstract}

\section{Introduction}
Video Domain Adaptation (VDA)~\cite{dasgupta2022overcoming, li2023source, zara2023unreasonable, wei2024unsupervised} aims to enhance the generalizability of the model, enabling it to be used for video-based tasks across various environments. 
Typically, VDA methods rely on sufficient target data to align domains by minimizing cross-domain distribution discrepancies. 
However, collecting a substantial amount of videos from the target domain can be costly or impractical in real-world applications. 
Therefore, Few-Shot Video Domain Adaptation (FSVDA)~\cite{xu2023augmenting, peng2023relamix} is proposed to achieve domain adaptation in a few-shot scenario, where only a very limited number of labeled videos are available in the target domain.

Existing FSVDA methods mainly rely on cross-domain feature alignment in the RGB image space. 
However, video data is always associated with multi-modal information (e.g. appearance, audio, and motion) which can provide complementary information to enhance the recognition procedure~\cite{Shahroudy2018MMaction, munro2020multi, Joze2020MMTM}. 
Consequently, there is a critical need for approaches to explore the multimodal nature of video within each domain. 
In this paper, we propose to study multimodal FSVDA, where both the source and target videos contain RGB and optical flow modalities.

\begin{figure}
 \centering
 \includegraphics[width=1.0\linewidth]{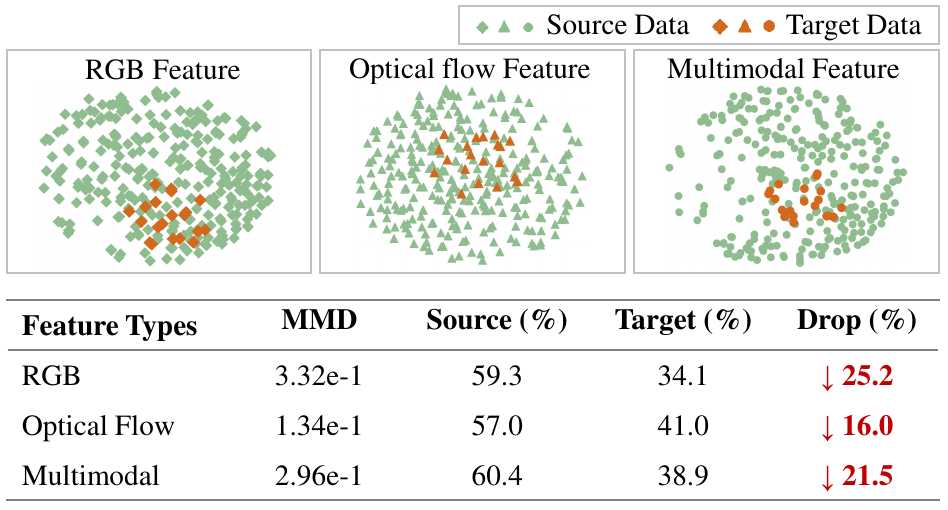}
 \caption
 {
 T-SNE visualization of multimodal features (from EPIC-Kitchens dataset~\cite{damen2018scaling}) in the FSVDA task, accompanied by the Maximum Mean Discrepancy (MMD) between the source and target domains, and the accuracy drop when the source model is directly applied to the target domain. 
 }
 \label{fig:motivation}
\end{figure}

It is significantly challenging to utilize multimodal features in FSVDA because modality collaboration and domain alignment are easily intertwined. 
Figure~\ref{fig:motivation} provides an example of the distribution discrepancy between the source and target domain, accompanied by the accuracy drop and maximum mean discrepancy. 
This figure highlights three important issues in the multimodal FSVDA task. 
\textbf{(1)}
Due to data scarcity, the feature distribution of the target instances often fails to accurately reflect the holistic distribution space of the target domain. 
Therefore, directly aligning the source and target distributions may deteriorate the generalizability of the source model. 
\textbf{(2)}
The distribution of RGB features shows a more obvious domain shift compared to optical flow features, indicating the need to treat individual unimodal features differently for domain alignment. 
\textbf{(3)}
Although fused multimodal features perform better than the unimodal features on the source domain due to the modality complementation, they yield worse results than the unimodal features on the target domain.
This indicates that the performance advantage brought by exhaustive modality combination is easily counteracted by the imbalance of the domain shifts in different modalities. 
Therefore, to utilize multimodal features more effectively, we need to carefully identify the common features across different modalities that lie in the same level of domain shift. 
Although existing multimodal domain adaptation~\cite{munro2020multi, lv2021differentiated, kim2021learning, song2021spatio, yang2022interact, wei2024unsupervised} and FSVDA~\cite{motiian2017few, li2021supervised, xu2023augmenting, peng2023relamix} methods address one of the three issues, none simultaneously considers all of them.

\begin{figure}[t]
 \centering
 \includegraphics[width=1.0\linewidth]{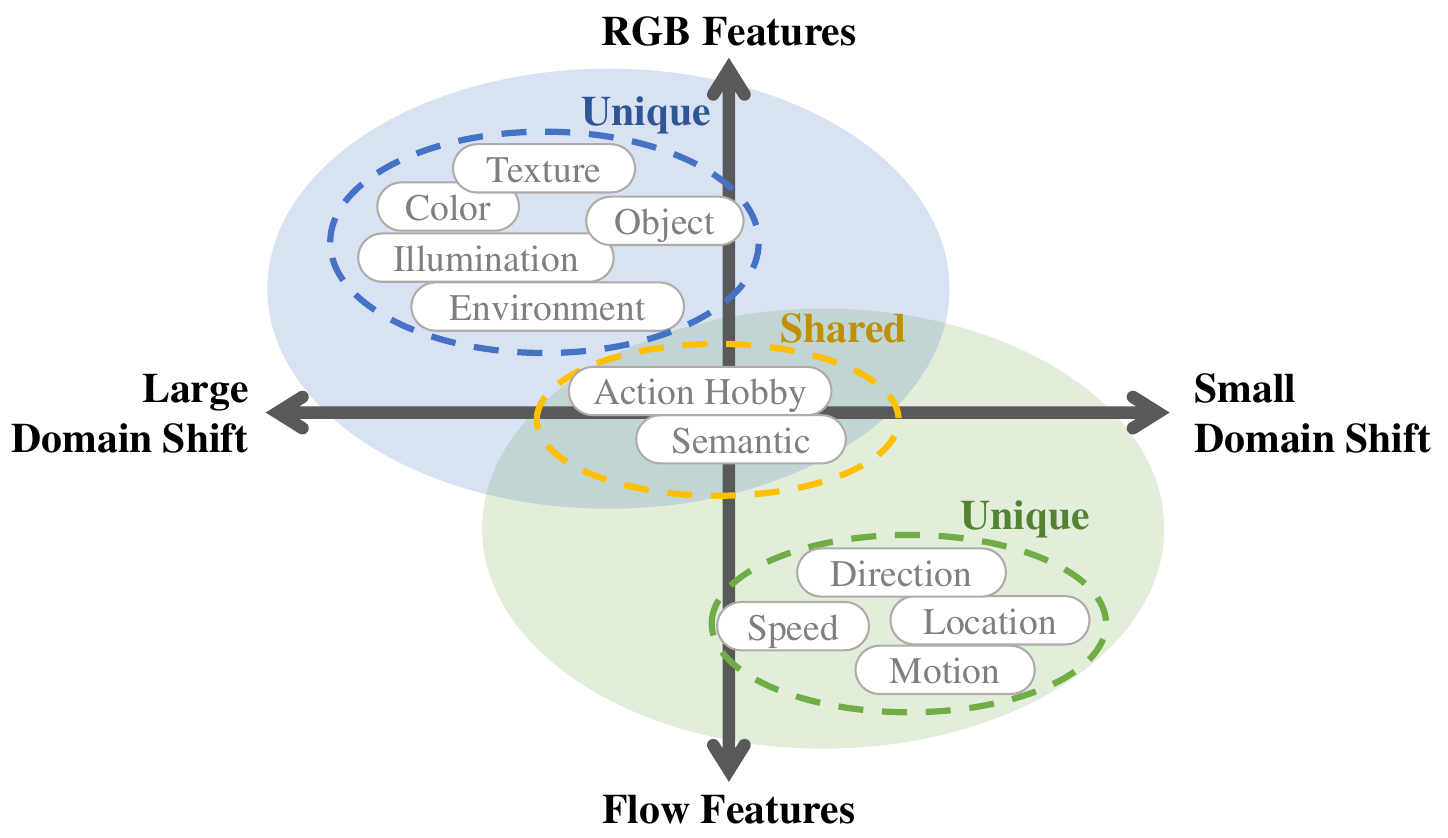}
 \caption
 { \color{black}
 Illustration of multimodal feature distribution, where each modality consists of distinct components. 
 Modality-unique features include, for instance, color and texture in RGB, and motion direction in optical flow, while modality-shared features exist across modalities at similar domain shift levels. 
 }
 \label{fig:motivation2}
\end{figure}

In this paper, we propose a new framework of Modality-Collaborative Low-Rank Decomposers (MC-LRD) to comprehensively address the three issues. 
The main idea of MC-LRD lies in efficiently learning modality-unique and modality-shared features that are more friendly for few-shot domain adaptation. 
{\color{black}
As the example illustrated in Figure~\ref{fig:motivation2}, each modality exhibits multiple conceptual components, each subject to varying degrees of domain shift, which present different levels of adaptation difficulty. 
These components can be classified into modality-unique and modality-shared properties (Figure~\ref{fig:motivation2}). 
Specifically, modality-unique features contain information specific to each modality, such as color, texture, and background context in RGB, or motion direction and speed in optical flow. 
In contrast, modality-shared features contain information shared across modalities and located at the same level of the domain shift. 
If features with differing domain shifts are treated equally in domain adaptation, components with larger domain shifts may remain under-adapted (\textit{limiting generalization}), while those with smaller shifts may become over-adapted (\textit{losing discriminability}). 
By decomposing modality-specific and modality-shared components from unimodal features, our approach enables targeted domain alignment tailored to the specific level of domain shift experienced by different features. 
}

To achieve this goal, the proposed MC-LRD is optimized with a two-stage training paradigm. 
In the \textbf{pre-training step}, we utilize a sufficient number of labeled source videos to optimize the base model. 
In the \textbf{adaptation step}, we extend the base model with modality-collaborative low-rank orthogonal decomposers to efficiently produce multimodal features that are more amenable to domain alignment. 
To comprehensively consider the short-term and long-term sequential features of the RGB and optical flow data, the MC-LRD is conducted at clip-level and video-level respectively. 
In either clip-level or video-level, the MC-LRD is comprised of multiple decomposers for each modality instantiated with low-rank orthogonal decomposers. 
{\color{black}
The progressively shared design facilitates mutual guidance in feature decomposition by establishing incremental cross-modal connections. 
}
The orthogonal constraint is adopted to minimize the interdependence of different decomposers and ensure that each decomposer can capture a dedicated aspect of the feature characteristics in multimodal feature decomposition. 
Meanwhile, Multimodal Decomposition Routers (MDR) are adopted to select appropriate decomposers for each input sample to produce modality-shared features and modality-unique features for domain alignment. 
The MDR consists of three sub-routers, i.e., one modality-shared sub-router for learning modality-shared features by leveraging cross-modal correlations with consistent activation weights on the RGB decomposers and the optical flow decomposers, and two modality-unique sub-routers for learning modality-unique features of the RGB and optical flow respectively. 
Furthermore, we propose a cross-domain activation consistency loss to ensure that target and source samples of the same category exhibit consistent activation preferences of the decomposers, thereby facilitating domain alignment.

Our main contributions are summarized as follows: 
\begin{enumerate}
 \item 
 We propose a novel framework of modality-collaborative low-rank decomposers, which is adept at handling the intricate interplay of modality collaboration and domain alignment in the underexplored task of multimodal few-shot video domain adaptation. 
 \item 
  To achieve efficient decomposition and domain adaptation, 
  we design the {\color{black} progressively shared} decomposers to more effectively capture distinct feature characteristics in multimodal feature decomposition. In addition, a cross-domain activation consistency loss is proposed to ensure that the decomposed multimodal features are more conducive to domain alignment. 
 \item 
 We evaluate the proposed method on three benchmark datasets and demonstrate its effectiveness with extensive experimental results. 
\end{enumerate}

\section{Related Work}

\subsection{Unsupervised Video Domain Adaptation}
Unsupervised video domain adaptation~\cite{choi2020shuffle, da2022dual, dasgupta2022overcoming, li2023source, lin2022cycda, zara2023unreasonable, wei2024unsupervised}, aims to recognize actions within a target domain by leveraging a model trained solely on annotations originating from out-of-domain source data. 
It presents challenges compared to the extensively studied image-based UDA~\cite{kang2019contrastive}, primarily due to the inherent complex modalities of video data. 
It also opens up opportunities to leverage multi-modal inputs for improved adaptation. 
MM-SADA~\cite{munro2020multi} firstly proposes exploring the multi-modal nature of videos for UDA, and leverages self-supervised alignment based on the correspondence of different modalities, in addition to adversarial alignment.
Song et al.~\cite{song2021spatio} and Kim et al.~.~\cite{kim2021learning} focus on integrating cross-modal information with contrastive learning, and incorporating both spatial and temporal aspects to improve domain alignment.
DLMM~\cite{lv2021differentiated} organizes an asynchronous learning group of the sub-models of different modalities for incremental optimization to deal with diverse domain shifts in different modalities.
MTRAN~\cite{huang2022relative} imitates domain shifts in multimodal and temporal dynamics by dividing target videos into source-like and target-like splits based on self-entropy, then employs a self-entropy-guided MixUp strategy to create synthetic samples, aligning them with hypothetical samples through multimodal and temporal relative alignment schemes. 
MD-DMD~\cite{yin2022mix} dynamically measures the adaptability score of each modality which enables modalities to teach each other domain adaptable knowledge by knowledge distillation. 
CIA~\cite{yang2022interact} enhances cross-domain alignment by using cross-modal interaction, allowing different modalities to share transferable information.
A3R~\cite{zhang2022audio} addresses domain shifts caused by changes in scenery by using activity sounds. 
Although MMVUDA methods enhance video model robustness, they necessitate a substantial amount of target domain data, which can be impractical in real-world applications. 

\subsection{Few-Shot Domain Adaptation }
(FSDA)~\cite{motiian2017few, xu2019d, gao2020pairwise1, gao2020pairwise2} addresses the challenge faced by traditional domain adaptation~\cite{lin2022cycda, zara2023unreasonable, wei2024unsupervised,10.1109/TMM.2023.3321430, 10.1109/TMM.2024.3361729}, which requires a significant amount of target domain data. 
It achieves domain generalization using only a few labeled target samples. 
FADA~\cite{motiian2017few} firstly explores domain adaptation in a few-shot scenario. 
It learns an embedded subspace with adversarial learning that aligns semantics between domains, allowing effective adaptation even with a few labeled target samples. 
d-SNE~\cite{xu2019d} uses stochastic neighborhood embedding and a modified-Hausdorff distance to mitigate domain shift problem for FSDA. 
PASTN~\cite{gao2020pairwise1} and PTC~\cite{gao2020pairwise2} employ adversarial learning and feature alignment techniques to address complex relationships between source and target domains for robust FSDA in the video domain.
Recently, SSA$^2$lign~\cite{xu2023augmenting} focuses on augmenting and attentively aligning snippet-level features through both semantic and statistical alignments. 
RelaMix~\cite{peng2023relamix} combines a temporal relational attention network with a latent space feature-mixing strategy to improve temporal generalizability and augment the shared latent space. 
{\color{black}
Cross-Domain Few-Shot Action Recognition (CDFSAR) \cite{wang2023cross, samarasinghe2023cdfsl, li2022cross, zhao2021domain, gao2022acrofod} is a task that bears close relevance to the task undertaken in this paper. 
Recent CSFSAR methods~\cite{markham2024understanding, wang2024tamt, guo2025dmsd} employ innovative strategies like data integration and hierarchical tuning to tackle domain shift challenges, minimizing the need for extensive retraining. 
This task specifically tackles few-shot video classification scenarios in which the seen and novel videos originate from distinct domains. 
Conversely, FSVDA emphasizes the adaptation process in scenarios where the target domain is represented by a minimal number of samples. 
}
However, these studies ignore the multimodal nature of videos. 
Although several methods have considered the multimodal nature of videos in unsupervised domain adaptation~\cite{munro2020multi, song2021spatio, kim2021learning, lv2021differentiated, yang2022interact, huang2022relative, yin2022mix, zhang2022audio}, they overlook considering both domain alignment and modality collaboration in a few-shot scenario. 

\subsection{Mixture of Experts}
(MoE)~\cite{jacobs1991adaptive, jordan1994hierarchical} consist of multiple experts and a gate network, outputting the weighted sum of the experts, with the gate values determined by the gate network on a per-example basis. 
Recently, sparse MoE has been widely used in natural language processing~\cite{shazeer2017outrageously, fedus2022switch, du2022glam}, computer vision~\cite{riquelme2021scaling} and multimodal learning~\cite{10.1109/TMM.2012.2229263, 10.1109/TMM.2024.3384058, 10.1109/TMM.2024.3399468} and shown remarkable achievements in fine-tuning large models. 
The sparse mixture of expert architectures scales model capacity without large increases in training or inference costs. 
Some recent works~\cite{wu2023mole,chen2024llava} propose to combine MoE and Low-Rank Adaptation (LoRA)~\cite{hu2021lora} as the Mixture of LoRA, offering flexibility in adapting to different requirements while minimizing computational overhead. 
Unlike existing MoE, our approach selectively activates LoRA decomposers to extract modality-unique and modality-shared features from video data that are more amenable to domain adaptation. 

\subsection{Decomposed Representation Learning }
(DRL) focuses on extracting underlying factors from observable data for meaningful representations. 
In recent years, it has seen significant advancements and has found applications in various domains~\cite{liu2021smoothing, cheng2020improving, zhang2020content, wang2020disentangled, 10.1109/TMM.2023.3347645, 10.1109/TMM.2023.3265843, 10.1109/TMM.2024.3360710}. 
Among applications of DRL, domain adaptation and multimodal feature representation learning are especially pertinent to our work. 
For domain adaptation, it is typically used to decompose domain-specific and domain-general factors to facilitate domain alignment~\cite{gonzalez2018image, lee2021dranet, kim2022learning, wei2024unsupervised}. 
Furthermore, DRL has been utilized in various multimodal tasks, including text-visual feature representation~\cite{tsai2018learning, alaniz2022compositional}, emotion recognition~\cite{li2023revisiting, yang2022disentangled}, and cross-modal generation~\cite{shi2019variational}. 
{\color{black}
For example, LEAD~\cite{qu2024lead} decouples features into source-known and -unknown components to identify target-private data to resolve the source-free universal domain adaptation task. 
However, this method is designed for unimodal domain adaptation and ignores the relationships between modalities for multi-modal domain adaptation. 
Different from existing DRL methods, our approach employs LoRA decomposers and multimodal decomposition routers to decompose features tailored for FSVDA, within a multimodal collaborative framework.
}

\begin{figure*}[t]
 \centering
 \includegraphics[width=\linewidth]{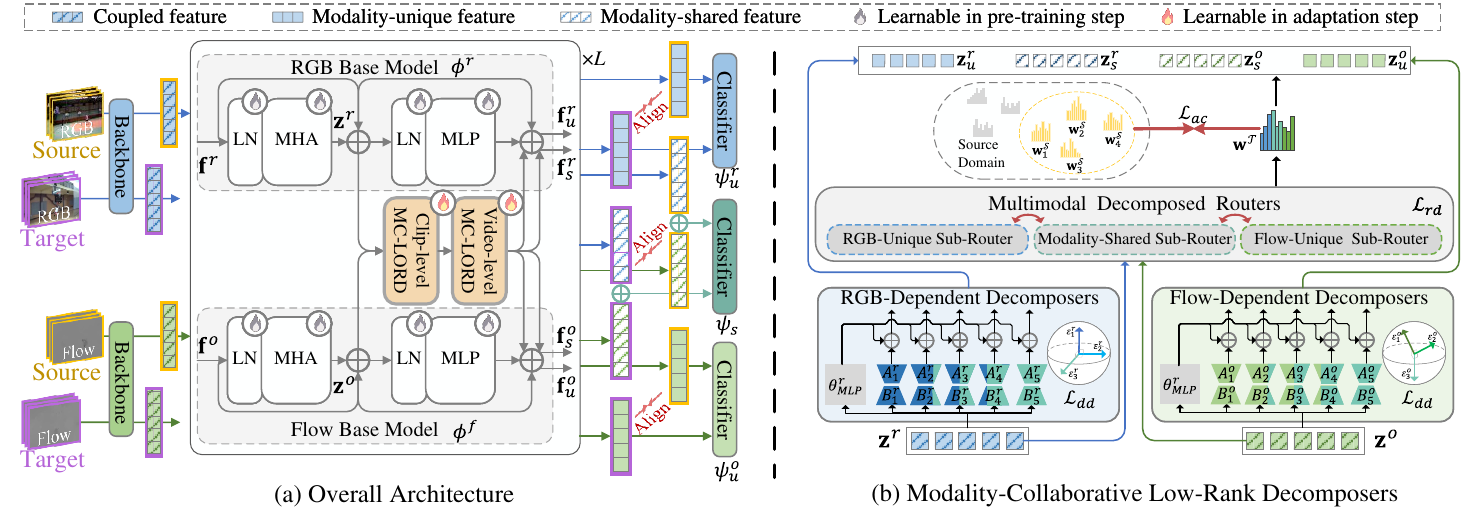}
 \caption{\color{black}
 (a) The MC-LRD framework consists of two stages: pre-training and adaptation. In pre-training, base models are trained using source videos to learn domain-specific knowledge. During adaptation, the Modality-Collaborative Low-Rank Decomposers (MC-LRD) use both source and target data to address domain shifts by refining multimodal features.
(b) MC-LRD takes multimodal features ($\mathbf{f}^r,\mathbf{f}^o$) as inputs, outputs disentangled features ($\mathbf{f}^r_u,\mathbf{f}^r_s,\mathbf{f}^o_u,\mathbf{f}^o_s$). 
 }
 \label{fig:framework}
\end{figure*}

\section{Methodology}
\subsection{Problem Definition}
In this paper, we provide formal definitions of multimodal Few-Shot Domain Adaptation (FSVDA) which is much more challenging than conventional domain adaptation, since only a few labeled videos are available on the target domain $\mathcal{T}$.
We assume that both source domain $\mathcal{S}$ and target domain $\mathcal{T}$ have multimodal data (i.e., RGB and optical flow in this work), where $\mathcal{S}$ and $\mathcal{T}$ exhibit distinct distributions but share the same label space. 
The domain $\mathcal{S}$ contains sufficient labeled samples $\mathbb{D}_\mathcal{S}=\{(\textbf{x}^r_i, \textbf{x}^o_i, y_i)\}_{i=1}^{N_\mathcal{S}}$, where $\textbf{x}_i^r=({x}_{i, 1}^r,{x}_{i, 2}^r,...,{x}_{i, T}^r)$ and $\textbf{x}_i^o=({x}_{i, 1}^o,{x}_{i, 2}^o,...,{x}_{i, T}^o)$ denote the instances for two modalities of the $i^{th}$ sample $\textbf{x}_i$, and $y_i \in \{1, 2, ..., \mathcal{C}\}$ denotes the class label. 
Target domain only contains $k$ instances for each class, denoted as $\mathbb{D}^{train}_\mathcal{T}=\{(\textbf{x}^r_i, \textbf{x}^o_i, y_i)\}_{i=1}^{N_\mathcal{T}^{train}}$, where ${N_\mathcal{T}^{train}}=\mathcal{C}\times k$. 
The objective is to utilize both $\mathbb{D}_\mathcal{S}$ and $\mathbb{D}^{train}_\mathcal{T}$ to train a model that generalizes well on the test set in target domain, denoted as $\mathbb{D}^{test}_\mathcal{T}=\{(\textbf{x}^r_i, \textbf{x}^o_i)\}_{i=1}^{N_\mathcal{T}^{test}}$, where $\mathbb{D}^{train}_\mathcal{T} \cap \mathbb{D}^{test}_\mathcal{T} = \emptyset$.

\subsection{Overall Architecture}
To solve the multimodal FSVDA task, we propose a new framework of Modality-Collaborative Low-Rank Decomposers.
The overall framework is designed as a two-stage training paradigm. 
In the \textbf{pre-training step}, we use sufficient source videos to train base models ($\phi^r, \phi^o$) for learning the source domain knowledge. 
In the \textbf{adaptation step}, we freeze parameters in base models and train the MC-LRD with both source data and target data to eliminate the domain shift among modality-unique and modality-shared features. 
Figure~\ref{fig:framework}(a) illustrates the pipeline of the adaptation step.

More specifically, we first encode each input video $(\textbf{x}^r, \textbf{x}^o)$ into feature representations $\mathbf{f}^r, \mathbf{f}^o$ with the backbone network. 
The original unimodal feature $\mathbf{f}^m=[f^m_{1},f^m_{2}, ..., f^m_{T}]^\top \in \mathbb{R}^{T \times d_{in}}$ is a sequence of clip-level features, where $m\in\{r, o\}$ represents the modality of RGB or optical flow, and $T$ is the number of clips. 
Then, we utilize an $L$-layer Transformer encoder to build the base model for each modality, defined as $\phi^r$ or $\phi^o$. 
Each layer in the Transformer encoder comprises a multi-head self-attention mechanism (MSA) and a feed-forward neural network (MLP) with parameters denoted as $\theta_{MLP}^m$: 
%
\begin{align}
\label{eq: Transformer-MSA}
\mathbf{z}^{m}_\ell &=\mathrm{MSA}(\mathrm{LN}(\mathbf{f}^m_{\ell-1}))+\mathbf{f}^m_{\ell-1}, &\ell=1...L,\\
\label{eq: Transformer-MLP}
\mathbf{f}^m_{\ell} &=\mathrm{MLP}(\mathrm{LN}(\mathbf{z}^{m}_\ell))+\mathbf{z}^{m}_\ell, &\ell=1...L,
\end{align}
%
%
where $\mathrm{LN}$ denotes the liner normalization operation, $\mathbf{f}^m_{0}$ is initialized as $\mathbf{f}^m$, and $\mathbf{z}^{m}_\ell \in \mathbb{R}^{T \times d}$.
%
%

{\color{black}
To learn more appropriate multimodal features for domain alignment, we further expand base models with multiple low-rank decomposers to adjust the MLP of the base models, resulting in decomposing modality-unique and modality-shared components. 
}
The MC-LRD adopts a multi-scale design at both clip-level and video-level, enabling it to capture short-term and long-term sequential features in multimodal videos. 
At both levels, the MC-LRD employs multiple modality-dependent decomposers, each instantiated with mutually orthogonal low-rank decomposers to extract distinct aspects of the multimodal features. 
{\color{black}
The Multimodal Decomposition Routers (MDR) module outputs different weights for different decomposed components to selectively combine outputs from these decomposers to produce modality-shared and modality-unique features. 
}
The learning processes for both decomposers and routers are constrained by orthogonal decorrelation losses to ensure efficient decomposition. 
Furthermore, a cross-domain activation consistency loss is leveraged to ensure that target and source samples of the same category exhibit consistent activation preferences of the decomposers, thereby facilitating domain alignment. 
After multi-layer iteration, we ultimately obtain modality-unique features $\mathbf{f}^m_{u,L}$ for each modality along with the modality-shared feature that is computed as the average of the modality-shared features $\mathbf{f}^m_{s,L}$ of each modality. 
These features are utilized to individually learn classifiers ($\psi^r_{u}, \psi^o_{u}, \psi_{s}$), and the final result is obtained by aggregating the outputs from these classifiers.

\subsection{Modality-Collaborative Low-Rank Decomposers}
In this section, we introduce the multi-scale framework, MC-LRD, which is designed to consider both clip-level and video-level features.
We detail the decomposers and multimodal decomposition routers in MC-LRD, highlighting the proposed orthogonal decorrelation constraints and cross-domain activation consistency loss,
which are beneficial for efficient decomposition and domain adaptation. 
The network structure is illustrated in Figure~\ref{fig:framework}(b).
\label{sec:MC-LRD}

\subsubsection{Clip-level Low-Rank Decomposers}

{
\color{black}
We extend LoRA \cite{hu2021lora} to construct 
$N_c$ low-rank decomposers adapting the MLP parameters $\theta^m_{MLP}$ for calculating decomposed components for input features. 
These decomposed components represent distinct aspects of the multimodal features, each exhibiting varying levels of domain shift. 
}
The decomposers can be formulated as:

\begin{equation}
  \begin{aligned} 
\label{eq:clip-level decomposers} 
 \mathcal{E}^{r}_{c,i}(\mathbf{z}^{r})
=&
\mathbf{z}^{r}(\alpha_c B^{r}_{c}+(1-\alpha_c)\hat{B}_{c})(\alpha_c A^{r}_{c}+(1-\alpha_c)\hat{A}_{c}) 
\\&+ \mathrm{MLP}(\mathbf{z}^{r}), 
\end{aligned}
\end{equation}
\begin{equation}
  \begin{aligned}
 \mathcal{E}^{o}_{c,i}(\mathbf{z}^{o})
=&
\mathbf{z}^{o}(\alpha_c B^{o}_{c}+(1-\alpha_c)\hat{B}_{c})(\alpha_c A^{o}_{c}+(1-\alpha_c)\hat{A}_{c}) 
\\& + \mathrm{MLP}(\mathbf{z}^{o}),
\end{aligned}
\end{equation}
where $\mathcal{E}^{r}_{c,i}$ and $\mathcal{E}^{o}_{c,i}$ handles the sequence of clip-level features,
$\mathbf{z}^{r}$ and $\mathbf{z}^{o}$ is defined in Eq(\ref{eq: Transformer-MSA}). 
We denote the corresponding MLP operation in Eq(\ref{eq: Transformer-MLP}) as $\mathrm{MLP}(\cdot)$. 
{\color{black}
We contend that distinct decomposers are intentionally designed to extract different types of features. 
To maximize the use of knowledge from the other modality, progressively shared parameters are assigned to decomposers of each modality.
This progressive design allows each decomposer to be influenced by varying degrees of cross-modal information, enabling a focus on distinct feature components. 
As illustrated in Figure~\ref{fig:framework}(b), this interaction is quantitatively expressed through the aggregation of $A^{m}_{c}$,$B^{m}_{c}$ and $\hat{A}_{c}$,$\hat{B}_{c}$, where $m\in{r,o}$, and the aggregation weight $\alpha_c$ is set as $\frac{N_c-i}{N_c-1}$. 
$A^{m}_{c}\in \mathbb{R}^{d_{ra}\times d}$ and $B^{m}_{c}\in \mathbb{R}^{d\times d_{ra}}$ are trainable low-rank matrix components specific to each modality, with $d_{ra}$ denoting their shared rank. 
In contrast, $\hat{A}_{c} \in \mathbb{R}^{d_{ra} \times d}$ and $\hat{B}_{c} \in \mathbb{R}^{d \times d_{ra}}$ are jointly learned low-rank components across modalities. 
The decomposers of each modality progressively share parameters with the other modality, transitioning from fully independent parameters ($A^{m}_{c}$,$B^{m}_{c}$) in the first pair of decomposers ($i=1$) to fully shared parameters($\hat{A}_{c}$,$\hat{B}_{c}$) in the last ($i=N_c$) pair of decomposers. 
The progressively shared design enables effective feature decomposition by establishing incremental cross-modal connections, allowing knowledge from each modality to guide decomposers to extract modality-unique and modality-shared features. 
}

To ensure each decomposer can capture an exclusive aspect of the multimodal features, we propose a {\textit{decomposer decorrelation loss}}, which is formulated as: 
\begin{equation}
\label{eq:loss_od}
\mathcal{L}_{dd} = 
\sum_{m\in\{r,o\}}
\sum_{i=1}^{N_c} \sum_{j=i+1}^{N_c}
\frac{ \langle \mathcal{E}^{m}_{c,i}(\mathbf{z}^{m}), \mathcal{E}^{m}_{c,j}(\mathbf{z}^{m}) \rangle }{\|\mathcal{E}^{m}_{c,i}(\mathbf{z}^{m})\| \|\mathcal{E}^{m}_{c,j}(\mathbf{z}^{m})\|},
\end{equation}
where $\mathcal{E}_{c,i}^m$ denotes the $i^{th}$ decomposer, and $\langle \cdot, \cdot \rangle$ means the inner product. 
{\color{black}
Minimizing this term ensures that the decomposer outputs are pairwise orthogonal, resulting in exclusive aspects of the multimodal features. 
The decomposed components constitute modality-specific and modality-shared features, with each component reflecting different degrees of domain shift.
}

Furthermore, we propose to obtain modality-unique and modality-shared features by applying different weights to activate the outputs of decomposers. 
{\color{black}
Therefore, we design the Multimodal Decomposition Routers (MDR) module to estimate the soft-merging weights for different decomposers: }
\begin{equation}
  {\mathbf{w}} = [w_u^r; w_u^o; w_s] = \mathrm{MDR}(z^r, z^o). 
\end{equation}
Specifically, modality-unique weights ${w}^r_{u}$, ${w}^o_{u}\in \mathbb{R}^{N_c}$  for RGB and optical flow modalities are calculated by modality-unique sub-routers $R_{u}^r$, $R_{u}^o$, and the modality-shared weight ${w}_{s}\in \mathbb{R}^{N_c}$ is produced by modality-shared sub-router $R_{s}$. 
$R_{u}^m$ and $R_{s}$ are formulated as single fully connected layers. 
By implementing a weight-sharing strategy, ${w}_{s}$ depends on a single router $R_{s}$ rather than training unimodal-unique routers, thereby capitalizing on cross-modal correlations. 
The calculation can be formulated as: 
\begin{align}
  \label{eq: weight unique}
& {w}^m_{u} = softmax\left(R_{u}^m(\mathrm{TAP}(\mathbf{z}^{m}))\right), \quad m \in \{r,o\}, \\
 \label{eq: weight shared}
& {w}_{s} = softmax\left(R_{s}(\mathrm{TAP}([\mathbf{z}^r; \mathbf{z}^{o}]))\right), 
\end{align}
where $\mathrm{TAP}(\cdot)$ denotes the temporal average pooling operation.

{\color{black}
With modality-unique weights ${w}^r_{u}$, ${w}^o_{u}$ and modality-shared weight ${w}_s$, decomposed modality-unique outputs $\mathbf{z}^{m}_{u}\in\mathbb{R}^{T\times d}$ and modality-shared outputs $\mathbf{z}^m_s\in\mathbb{R}^{T\times d}$ are computed with the activation rules as: }
%
%
%
\begin{align}
\label{eq: early_disentangle_unique}
  & \mathbf{z}^{r}_{u}=\sum_{i=1}^{N_c}{w}^r_{u,i}\cdot \mathcal{E}_{c,i}^r(\mathbf{z}^{r}), \quad \mathbf{z}^{o}_{u}=\sum_{i=1}^{N_c}{w}^o_{u,i}\cdot \mathcal{E}_{c,i}^o(\mathbf{z}^{o}), \\
\label{eq: early_disentangle_shared}
  &\mathbf{z}^{r}_{s}=\sum_{i=1}^{N_c}{w}_{s,i}\cdot 
  \mathcal{E}_{c,i}^r(\mathbf{z}^{r}), \quad
  \mathbf{z}^{o}_{s}=\sum_{i=1}^{N_c}{w}_{s,i}\cdot \mathcal{E}_{c,i}^o(\mathbf{z}^{o}). 
\end{align}
%
%
%
{\color{black}
Modality-unique weights emphasize features specific to each modality, while shared weights capture commonalities across modalities.  These weights are applied to encoded representations to selectively extract and combine features, enabling the disentanglement of unique and shared aspects. 
}

To ensure that different decomposers are distinctly activated to learn either modality-unique or shared features, we introduce a \textit{router decorrelation loss}, which enforces orthogonality among the outputs of the modality-unique sub-routers and the modality-shared sub-router: 
\begin{equation}
\label{eq:loss_rd}
 \mathcal{L}_{rd} = \sum_{m\in\{r,o\}} 
 \langle {w}^m_{u}, {w}_{s} \rangle.
\end{equation}

Then, we further introduce a {\textit{cross-domain activation consistency loss}} to ensure that samples in the target and source domains share the same activation preferences, thereby enforcing cross-domain consistency. 
Specifically, we minimize the divergence between the activation weights of the target domain sample and the average activation weights of source domain samples that belong to the same class as the current target sample. 
The loss function is defined as follows: 
\begin{equation}
\label{eq:loss_ac}
  \mathcal{L}_{\mathrm{ac}}=
  \left\|
  {\mathbf{w}}^\mathcal{T}-
  \frac{1}{|\mathbb{D}_\mathcal{S}^{y^\mathcal{T}}|}
  \sum_{\mathbf{x}^\mathcal{S}\in\mathbb{D}_\mathcal{S}^{y^\mathcal{T}}}
  {\mathbf{w}}^\mathcal{S}
  \right\|_2^2, 
\end{equation}
where ${\mathbf{w}}^\mathcal{T}$, ${\mathbf{w}}^\mathcal{S}$ represent the MDR output for the target sample $\mathbf{x}^\mathcal{T}$ or the source sample $\mathbf{x}^\mathcal{S}$. 
Additionally, $\mathbb{D}_\mathcal{S}^{y^\mathcal{T}}\subset \mathbb{D}_\mathcal{S}$ denotes the set of videos in the source domain that belong to the same class $y^\mathcal{T}$ as the target sample $\mathbf{x}^\mathcal{T}$.

\subsubsection{Video-level Low-Rank Decomposers}
Following clip-level decomposers, we explore diverse sequential features from a global perspective with video-level low-rank decomposers. 
{\color{black}
These decomposers integrate the long-term temporal context of input data, focusing on the motion of relevant objects in the video, and breaking down the temporally dependent components. 
}
For each modality, we instantiated $N_v$ video-level decomposers $\mathcal{E}^{r}_{v,i}$ and $\mathcal{E}^{o}_{v,i}$:

\begin{equation}
  \begin{aligned}
 \small 
\mathcal{E}^{r}_{v,i}(\mathbf{z}^{r})
= &
\mathbf{z}^{r\top}(\alpha_v B^{r}_{v}+(1-\alpha_v)\hat{B}_{v})(\alpha_v A^{r}_{v}
+(1-\alpha_v)\hat{A}_{v})^\top
\\& + \mathrm{MLP}(\mathbf{z}^{r}), 
\end{aligned}
\label{eq:video-level decomposers_rgb}
\end{equation}

\begin{equation}
  \begin{aligned}
\mathcal{E}^{o}_{v,i}(\mathbf{z}^{o})
= &
\mathbf{z}^{o\top}(\alpha_v B^{o}_{v}+(1-\alpha_v)\hat{B}_{v})(
\alpha_v A^{o}_{v}+(1-\alpha_v)\hat{A}_{v})^\top
\\& + \mathrm{MLP}(\mathbf{z}^{o}),
\end{aligned}
\label{eq:video-level decomposers_flow}
\end{equation}
where $\mathbf{z}^{m}_{u}$ denotes the preliminary decomposed features obtained from the clip-level decomposers, with the calculation process introduced in Eq(\ref{eq: early_disentangle_unique}). 
Similarly, the calculation for modality-shared feature $\mathbf{z}^{m}_{s}$, obtained from Eq(\ref{eq: early_disentangle_shared}), follows the same methodology as that in Eq(\ref{eq:video-level decomposers_rgb}) and Eq(\ref{eq:video-level decomposers_flow}). 
$\mathrm{MLP}(\cdot)$ denotes the corresponding MLP operation in Eq(\ref{eq: Transformer-MLP}). 
$A^{m}_{v}\in \mathbb{R}^{d_{ra}\times T}$, $B^{m}_{v}\in \mathbb{R}^{T\times d_{ra}}$, $\hat{A}_{c} \in \mathbb{R}^{d_{ra} \times d}$ and $\hat{B}_{c} \in \mathbb{R}^{d \times d_{ra}}$ are trainable low-rank matrix, and $\alpha_v=\frac{N_v-i}{N_v-1}$. 
{\color{black}
In contrast to Eq(\ref{eq:clip-level decomposers}), the subtle use of transposition ($\top$) enables LoRA decomposers to achieve global information exploration through the mixing of clip-level features. 
It enables the interaction of clip-level features and learning higher-order video-level dependencies with simple low-rank decomposers. 
}
In video-level decomposer, the Multimodal Decomposition Routers (MDR) module is also employed, which takes $\mathbf{z}^{r}_{u}$, $\mathbf{z}^{o}_{u}$ or $\mathbf{z}^{r}_{s}, \mathbf{z}^{o}_{s}$ as input to compute the activation weights for modality-unique $\mathbf{\hat{z}}^{m}_u$ and modality-shared features $\mathbf{\hat{z}}^{m}_s$ respectively.
The calculation is performed in the same manner as in Eq(5-9). 
In video-level low-rank decomposers, we apply $\mathcal{L}_{dd}$, $\mathcal{L}_{rd}$ and $\mathcal{L}_{ac}$ analogous to those in Eq(\ref{eq:loss_od}), Eq(\ref{eq:loss_rd}) and Eq(\ref{eq:loss_ac}) to enforce constraints on the decomposition process. 
The final output $\mathbf{{f}}^{m}_u$ and $\mathbf{{f}}^{m}_s$ are updated as $\mathbf{\hat{z}}^{m}_u$ and $\mathbf{\hat{z}}^{m}_s$. 

Note that, for simplicity, we describe the framework with a single-layer structure. 
In practical experiments, the second and subsequent layers of the multi-layer Transformer use the decomposed features $\mathbf{f}^{m}_{u, \ell-1}, \mathbf{f}^{m}_{s, \ell-1}$ as input for further decomposition.  
\subsection{Learning Objectives}
\label{sec:learning objectives}

In addition to the previously introduced constraints, we use the classification loss $\mathcal{L}_{cls}$ and adversarial domain alignment $\mathcal{L}_{ada}$ to optimize the model~\cite{wei2024unsupervised}. 
Combining these losses, the final objective function is formulated as follows: 
\begin{equation}
 \label{eq: objective loss}
\mathcal{L}=\mathcal{L}_{cls}+\hat{\mathcal{L}}_{dd}+\hat{\mathcal{L}}_{rd}+\hat{\mathcal{L}}_{ac}+\mathcal{L}_{ada}. 
\end{equation}
Here, $\hat{\mathcal{L}}_{dd}$, $\hat{\mathcal{L}}_{rd}$ and $\hat{\mathcal{L}}_{ac}$ indicate averaged corresponding losses at both the clip level and video level decomposers in all layers of MC-LE. 
To balance the contributions of the different loss functions, we follow~\cite{wei2024unsupervised} to empirically set all trade-off parameters to 1.

\section{Experiments}
\subsection{Experimental Setup}
\label{sec:experiment-setup}

\noindent
\subsubsection{Dataset}

\begin{itemize}
  \item 
  \textbf{EPIC-Kitchens}
is a challenging dataset, which consists of fine-grained daily activity videos collected from a first-person view in a kitchen scene~\cite{damen2018scaling}. 
Following~\cite{munro2020multi}, we conducted experiments on three domain partitions (D1, D2, and D3) of the 8 largest action classes. 
It contains 2495/313 train/test action videos on D1, 1543/417 train/test action videos on D2, and 3897/1030 on D3 train/test action videos. 
  \item 
  \textbf{UCF-HMDB}
is one of the most widely used cross-domain video datasets. 
It has 12 shared classes, respectively from UCF~\cite{soomro2012ucf} and HMDB~\cite{kuehne2011hmdb}. 
We follow the train/test split used in~\cite{yang2022interact, peng2023relamix}. 
%
It contains 3,209 videos in total with 1,438 training videos and 571 validation videos from UCF, and 840 training videos and 360 validation videos from HMDB. 
There are two settings of interest: UCF $\to$ HMDB (U$\to$H) and HMDB $\to$ UCF (H$\to$U). 
  \item 
  \textbf{Jester.}
is a large-scale dataset containing 148,092 video clips of people performing a variety of basic hand gestures~\cite{materzynska2019jester}. 
Following the cross-domain benchmark provided by~\cite{pan2020adversarial}, it contains 51,498 video clips for the training set and 51,415 video clips for the test set, covering seven distinct gesture classes.

\end{itemize}
\subsubsection{Baseline}
We mainly compared MC-LRD with several existing state-of-the-art methods and tasks. 
\begin{itemize}
  \item \textbf{Few-Shot Domain Adaptation (FSDA)} 
  (eg. SSA$^2$lign~\cite{xu2023augmenting} and RelaMix~\cite{peng2023relamix})
  Unlike our approach, these existing methods are primarily designed for unimodal video data. 
  \item \textbf{Unsupervised Domain Adaptation (UDA)} (eg. TranSVAE~\cite{wei2024unsupervised})
  aims to adapt models trained on labeled source domains to effectively perform on target domains experiencing domain shifts, leveraging sufficient unlabeled target samples. 
  \item \textbf{Few-Shot Action Recognition (FSAR)} (eg. TRX~\cite{perrett2021trx}, HyRSM~\cite{wang2022hyrsm})
  focus on identifying actions within videos using only a limited set of labeled instances per action class, allowing rapid learning in few-shot scenarios absent of domain shift.
\end{itemize}
{
For UDA and FSDA methods, we follow the reformulations in~\cite{peng2023relamix} to extend these methods to few-shot domain adaptation. 
}
Since existing methods still cannot be directly applied to multimodal FSVDA, we extend them to the multimodal scenario using early-fusion and late-fusion strategies for a fair comparison. 
Note that late-fusion achieves better performance in most settings, so we present the multimodal baseline results using the late-fusion strategy. 

\subsubsection{Implementation Details.}

\begin{itemize}
  \item 
  \textbf{Data Preparation }
In this study, we propose to investigate the multi-modal few-shot video domain adaptation task, thus necessitating the acquisition of multimodal video data. 
{\color{black}
EPIC-Kitchen~\cite{damen2018scaling} provided the official optical flow data. 
For the UCF~\cite{soomro2012ucf}, HMDB~\cite{kuehne2011hmdb}, and Jester~\cite{materzynska2019jester} datasets, we generate optical flow data from the raw video. 
Specifically, we utilize the dense optical flow algorithm~\cite{lucas1981iterative} to create an optical flow sequence for each video. 
}
\item 
\textbf{{Backbones.}}
We follow ~\cite{munro2020multi, wei2024unsupervised, peng2023relamix} to adopt the I3D~\cite{carreira2017quo} architecture with frozen weights pre-trained on Kinetics400~\cite{kay2017kinetics}, as the backbone feature extractors for all methods and experiments conducted in this section. 
The dimensions of clip-level features ($d_{in}$) extracted from I3D are 2048 for the RGB and optical flow modalities. 
Following~\cite{wei2024unsupervised, peng2023relamix}, we set the clip number $T$ to 12 for both RGB and optical flow modalities in each video. 
Similarly, we utilize the I3D architecture pre-trained on optical flow data to extract optical flow features. 
Specifically, we sample 16 frames along videos using a temporal window that slides with a stride of 1. 
For each clip, the temporal window includes the previous seven clips and the subsequent eight clips, with zero padding applied at the beginning and end of the video. 
These sliding windows are then input into the I3D backbone, producing a 2048-dimensional feature vector for each clip. 
\item 
\textbf{Network Architecture.}
We leverage the Transformer architecture~\cite{vaswani2017attention} as the modality-unique base model, where the hidden dimension, head number, and layer number are set to be 512, 6, and 2 respectively. 
In the Modality-Collaborative Low-Rank Decomposers, we set the number of decomposers $N_c$=$N_v$ to 6, and the rank $d_{ra}$ is set to 64. 
\item 
\textbf{Learning. }
Our MC-LRD is implemented with PyTorch~\cite{paszke2019pytorch}.
Our model and baselines are all trained with the Adam optimizer. 
In the pre-training step, we optimize base models with the classification constraint for 2 epochs, where the learning rate is $10^{-5}$. 
In the adaptation step, we optimize the parameters of the proposed MC-LRD with the objective defined in Eq(\ref{eq: objective loss}) for 50 epochs, where the learning rate is $10^{-4}$. 
The batch size is configured to 128 for the pretraining step (source videos) and the adaptation step (both source and target videos).
{\color{black}
We train our model on one NVIDIA RTX 4090 GPU for nearly 3 hours. 
}
\item 
\textbf{Evaluation.}
Refer to prior FSVDA works~\cite{peng2023relamix}, we randomly selected $k = (1, 5, 10, 20)$ labeled samples per class from target domain data to construct the target training set $\mathbb{D}^{train}_\mathcal{T}$. 
{
For EPIC-Kitchens and UCF-HMDB, we follow the benchmarks from~\cite{peng2023relamix}. 
For Jester, we establish the few-shot split using random selection. 
}
We apply such benchmarks to all the experiments in this section, facilitating fair comparisons. 
\end{itemize}

\begin{table*}[!t]
 \centering
\setlength{\tabcolsep}{4pt}
 \caption{Results on EPIC-Kitchens. `R' and `F' denote RGB and Optical Flow. The best results are presented in bold. }
\resizebox{\linewidth}{!}{
\begin{tabular}{c|l|cc|cc|cc|cc|cc|cc|cc}
\toprule
\multirow{2}{*}{Modality} & \multirow{2}{*}{Method} & \multicolumn{2}{c|}{D1$\to$D2} & \multicolumn{2}{c|}{D1$\to$D3} & \multicolumn{2}{c|}{D2$\to$D1} & \multicolumn{2}{c|}{D2$\to$D3} & \multicolumn{2}{c|}{D3$\to$D1} & \multicolumn{2}{c|}{D3$\to$D2} & \multicolumn{2}{c}{Mean}  \\
  & & 1-shot  & 5-shot  & 1-shot  & 5-shot  & 1-shot  & 5-shot  & 1-shot  & 5-shot  & 1-shot  & 5-shot  & 1-shot  & 5-shot  & 1-shot  & 5-shot  \\ \midrule \midrule
\multirow{7}{*}{R}
 &PASTN~\cite{gao2020pairwise} & 33.3 & 38.2 & 35.3 & 39.4 & 34.0 & 38.9 & 39.2 & 43.1 & 38.2 & 33.6  & 43.0 & 45.5 & 36.1 & 40.5 \\
 &TA$^3$N~\cite{chen2019temporal} & 36.8 & 39.0 & 36.7 & 40.2 & 36.8 & 38.9 & 41.1 & 43.4 & 33.1 & 40.0 & 42.8 & 45.8 & 37.9 & 41.2 \\
 & TRX~\cite{wei2024unsupervised} & 24.8 & 25.0 & 25.3 & 25.9 & 26.1 & 27.7 & 28.4 & 28.0 & 26.6 & 28.9 & 28.8 & 29.1 & 26.7 & 27.4 \\
 & HyRSM~\cite{wang2022hyrsm} & 31.1 & 33.5 & 33.2 & 37.2 & 33.4 & 32.7 & 40.4 & 40.3 & 35.0 & 34.8 & 41.6 & 41.8 & 35.8 & 36.7 \\
 & TranSVAE~\cite{wei2024unsupervised}  & 32.9 & 39.5 & 35.3 & 40.4 & 37.0 & 39.1 & 36.1 & 38.2 & 42.8 & 44.9 & 41.2 & 44.4 & 37.6 & 41.1 \\
 & SSA$^2$lign~\cite{xu2023augmenting} & 32.0 & 40.4 & 31.3 & 40.1 & 30.1 & 39.3 & 34.5 & 38.9 & 28.7 & 42.9 & 32.3 & 38.7 & 31.5 & 40.1 \\
 & RelaMix~\cite{peng2023relamix} & 39.1 & 43.9 & 38.4 & 41.6 & 38.4 & 42.1 & 37.9 & 41.6 & 45.1 & 46.2 & 45.5 & 48.0 & 40.7 & 43.9 \\ \midrule
\multirow{5}{*}{F}  
& TRX~\cite{wei2024unsupervised} & 22.1  & 23.5  & 23.6  & 23.9  & 24.2  & 26.7  & 24.5  & 25.3  & 25.4  & 25.8  & 25.2  & 26.7  & 24.2  & 25.3  \\
  & HyRSM~\cite{wang2022hyrsm} & 27.3  & 29.6  & 27.1  & 27.6  & 30.2  & 30.9  & 39.1  & 39.6  & 26.3  & 27.6  & 32.1  & 36.8  & 30.4  & 32.0  \\
  & TranSVAE~\cite{wei2024unsupervised}  & 33.2  & 40.4  & 34.3  & 34.4  & 36.3  & 39.3  & 44.5  & 41.8  & 35.9  & 41.6  & 46.9  & 48.3  & 38.5  & 41.0  \\
  & SSA$^2$lign~\cite{xu2023augmenting}  & 43.7  & 47.6  & 30.2  & 41.0  & 44.1  & 47.8  & 41.8  & 45.2  & 39.5  & 43.4  & 48.8  & 47.7  & 41.4  & 45.5  \\
  & RelaMix~\cite{peng2023relamix} & 43.9  & 48.9  & 32.5  & 40.2  & 44.1  & 38.3  & 43.7  & 41.2  & 40.1  & 43.0  & 49.6  & 53.3  & 42.3  & 44.2  \\ \midrule
\multirow{6}{*}{R+F}  & TRX~\cite{wei2024unsupervised} & 25.7  & 28.5  & 23.2  & 30.5  & 28.2  & 31.4  & 28.0  & 30.2  & 26.1  & 29.5  & 31.6  & 31.9  & 27.1  & 30.3  \\
  & HyRSM ~\cite{wang2022hyrsm} & 28.4  & 34.3  & 33.5  & 35.1  & 28.5  & 31.7  & 43.1  & 42.9  & 28.6  & 31.3  & 34.9  & 41.1  & 32.8  & 36.0  \\
  & TranSVAE~\cite{wei2024unsupervised}  & 38.9  & 44.7  & 34.9  & 40.0  & 43.0  & 42.5  & 47.1  & 50.1  & 36.6  & 44.8  & 52.0  & 54.0  & 42.1  & 46.0  \\
  & SSA$^2$lign~\cite{xu2023augmenting}  & 45.3  & 47.7  & 39.1  & 44.0  & 37.2  & 42.8  & 43.1  & 45.1  & 39.8  & 43.7  & 53.1  & 51.1  & 42.9  & 45.7  \\
  & RelaMix~\cite{peng2023relamix} & 48.1  & 46.0  & 39.8  & 44.1  & 41.5  & 44.8  & 48.1  & 50.4  & 42.4  & 45.9  & 51.6  & 55.6  & 45.7  & 47.2  \\
  & \textbf{Ours} & \textbf{51.9} & \textbf{53.5} & \textbf{44.9} & \textbf{46.1} & \textbf{46.8} & \textbf{49.4} & \textbf{50.8} & \textbf{53.3} & \textbf{49.1} & \textbf{52.2} & \textbf{55.7} & \textbf{58.7} & \textbf{49.9} & \textbf{52.2} \\ 
  \bottomrule
\end{tabular}
}
 \label{tab:comparation_Epic}
\end{table*}

\begin{table*}[ht]
\centering
\caption{Results on EPIC-Kitchens in the settings of 10-shot and 20-shot. }
\setlength{\tabcolsep}{3pt}
\resizebox{\linewidth}{!}{
\begin{tabular}{c|l|cc|cc|cc|cc|cc|cc|cc}
\toprule
\multirow{2}{*}{Modality} & \multirow{2}{*}{Method} & \multicolumn{2}{c|}{D1$\to$D2} & \multicolumn{2}{c|}{D1$\to$D3} & \multicolumn{2}{c|}{D2$\to$D1} & \multicolumn{2}{c|}{D2$\to$D3} & \multicolumn{2}{c|}{D3$\to$D1} & \multicolumn{2}{c|}{D3$\to$D2} & \multicolumn{2}{c}{Mean}  \\
  & & {10-shot} & {20-shot} & {10-shot} & {20-shot} & {10-shot} & {20-shot} & {10-shot} & {20-shot} & {10-shot} & {20-shot} & {10-shot} & {20-shot} & {10-shot} & {20-shot} \\ \midrule
\multirow{5}{*}{R}  & TRX & 25.2  & 25.9  & 28.1  & 28.8  & 30.7  & 31.6  & 30.6  & 31.9  & 29.3  & 30.0  & 28.4  & 33.1  & 28.7  & 30.2  \\
  & HyRSM & 34.0  & 37.2  & 36.5  & 36.7  & 33.9  & 34.8  & 41.2  & 41.4  & 35.7  & 35.0  & 41.5  & 41.5  & 37.1  & 37.8  \\
  & TranSVAE  & 39.5  & 42.8  & 37.5  & 41.7  & 40.3  & 42.3  & 37.5  & 41.4  & 44.5  & 45.9  & 45.6  & 45.6  & 40.8  & 43.3  \\
  & SSA$^2$lign & 37.6  & 41.5  & 40.5  & 41.6  & 42.0  & 42.6  & 41.1  & 39.1  & 42.1  & 44.5  & 41.9  & 42.7  & 40.9  & 42.0  \\
  & RelaMix & 43.7  & 47.9  & 42.1  & 42.8  & 42.5  & 43.1  & 42.3  & 42.5  & 47.4  & 46.5  & 48.1  & 48.1  & 44.4  & 45.2  \\ \midrule
\multirow{5}{*}{F}  & TRX & 25.4  & 26.6  & 25.4  & 26.9  & 28.6  & 30.1  & 26.8  & 28.3  & 28.2  & 29.5  & 26.8  & 30.5  & 26.9  & 28.7  \\
  & HyRSM & 31.4  & 33.8  & 28.1  & 28.7  & 31.8  & 31.7  & 40.4  & 40.9  & 29.8  & 30.7  & 38.2  & 39.5  & 33.3  & 34.2  \\
  & TranSVAE  & 42.1  & 40.9  & 38.3  & 36.3  & 40.7  & 40.7  & 44.6  & 42.6  & 40.0  & 42.8  & 47.3  & 48.0  & 42.2  & 41.9  \\
  & SSA$^2$lign & 50.8  & 49.6  & 43.4  & 43.1  & 41.8  & 42.3  & 46.4  & 47.8  & 44.6  & 50.0  & 50.9  & 50.8  & 46.3  & 47.3  \\
  & RelaMix & 46.7  & 50.6  & 43.2  & 42.8  & 42.5  & 44.3  & 47.1  & 45.7  & 44.2  & 50.0  & 51.5  & 55.8  & 45.9  & 48.2  \\ \midrule
\multirow{6}{*}{R+F}  & TRX & 29.7  & 32.7  & 29.7  & 29.2  & 32.7  & 35.7  & 32.8  & 29.4  & 31.5  & 28.9  & 32.1  & 32.9  & 31.4  & 31.5  \\
  & HyRSM & 34.7  & 35.7  & 34.1  & 37.1  & 32.8  & 33.3  & 43.5  & 44.5  & 34.4  & 36.3  & 39.7  & 40.4  & 36.5  & 37.9  \\
  & TranSVAE  & 44.0  & 47.1  & 41.5  & 38.9  & 42.8  & 44.8  & 47.6  & 48.2  & 43.9  & 43.7  & 53.9  & 54.7  & 45.6  & 46.2  \\
  & SSA$^2$lign & 50.5  & 52.9  & 45.6  & 46.2  & 44.1  & 46.2  & 51.3  & 51.3  & 48.0  & 50.5  & 55.1  & 52.4  & 49.1  & 49.9  \\
  & RelaMix & 47.5  & 54.7  & 46.3  & 46.0  & 46.6  & 48.0  & 50.6  & 50.7  & 44.9  & 50.5  & 54.1  & 56.0  & 48.3  & 51.0  \\
  & \textbf{Ours} & \textbf{54.0} & \textbf{56.5} & \textbf{48.8} & \textbf{49.5} & \textbf{49.6} & \textbf{52.8} & \textbf{55.2} & \textbf{56.6} & \textbf{52.2} & \textbf{52.7} & \textbf{58.3} & \textbf{58.4} & \textbf{53.0} & \textbf{54.4}  \\ \bottomrule
\end{tabular}}
\label{table:compare-epic-1020shot}
\end{table*}

\begin{table*}[t]
\centering
 \caption{Results on UCF-HMDB, and Jester datasets.}
\renewcommand{\arraystretch}{0.9}
\resizebox{1.0\linewidth}{!}{
\begin{tabular}{c|l|cccc|cccc|cccc}
\toprule
\multirow{2}{*}{Modality} & \multirow{2}{*}{Method} & \multicolumn{4}{c|}{U$\to$H} & \multicolumn{4}{c|}{H$\to$U} & \multicolumn{4}{c}{Jester} \\
& & 1-shot & 5-shot & 10-shot & 20-shot & 1-shot & 5-shot & 10-shot & 20-shot & 1-shot & 5-shot & 10-shot & 20-shot \\
\midrule \midrule
\multirow{5}{*}{R}
& TRX & 77.2 & 80.3 & 78.6 & 81.9 & 82.2 & 83.1 & 81.1 & 84.4 & 28.3 & 29.2 & 29.9 & 33.0 \\
& HyRSM & 79.7 & 81.1 & 82.2 & 83.6 & 88.1 & 90.1 & 91.0 & 90.8 & 30.5 & 31.8 & 31.9 & 34.8 \\
& TranSVAE & 75.3 & 79.2 & 83.2 & 84.8 & 62.3 & 75.0 & 94.4 & 95.1 & 35.2 & 36.3 & 35.0 & 35.4 \\
& SSA$^2$lign & 81.1 & 88.1 & 88.3 & 87.8 & 91.8 & 95.1 & 88.3 & 87.8 & 42.9 & 43.8 & 44.7 & 48.4 \\
& RelaMix & 85.6 & 91.1 & 91.1 & 92.2 & 94.1 & 97.2 & 97.9 & 98.4 & 43.8 & 46.8 & 47.5 & 47.7 \\ \midrule
\multirow{5}{*}{F}
& TRX & 72.3 & 74.6 & 75.8 & 77.6 & 80.1 & 81.8 & 81.7 & 82.3 & 25.8 & 26.2 & 28.5 & 30.8 \\
& HyRSM & 76.4 & 78.4 & 80.9 & 80.3 & 86.3 & 87.8 & 87.9 & 89.1 & 27.6 & 27.9 & 29.3 & 30.2 \\
& TranSVAE & 63.1 & 69.2 & 64.7 & 70.3 & 55.2 & 72.3 & 69.2 & 71.6 & 39.8 & 39.1 & 40.0 & 39.7 \\
& SSA$^2$lign & 78.1 & 83.3 & 87.8 & 88.9 & 83.0 & 93.5 & 95.6 & 97.7 & 41.5 & 49.2 & 50.3 & 53.1 \\
& RelaMix & 73.5 & 81.9 & 83.8 & 90.3 & 90.8 & 93.7 & 94.8 & 94.8 & 40.4 & 46.2 & 52.1 & 55.9 \\ \midrule
\multirow{6}{*}{R+F}
& TRX & 75.0 & 80.5 & 80.2 & 81.3 & 82.2 & 82.8 & 81.3 & 85.1 & 28.9 & 30.1 & 30.9 & 35.1 \\
& HyRSM & 81.6 & 82.3 & 83.1 & 82.8 & 89.1 & 89.9 & 91.0 & 92.3 & 31.1 & 31.9 & 33.8 & 37.0 \\
& TranSVAE & 77.5 & 85.6 & 81.1 & 81.9 & 69.0 & 84.8 & 81.1 & 86.2 & 42.9 & 41.7 & 43.1 & 42.1 \\
& SSA$^2$lign & 85.3 & 91.7 & 93.3 & 94.2 & 93.2 & 97.2 & 99.3 & 99.5 & 45.9 & 51.0 & 53.6 & 55.8 \\
& RelaMix & 85.1 & 90.3 & 91.1 & 94.0 & 94.6 & 97.4 & 98.3 & 98.4 & 47.0 & 51.0 & 55.4 & 58.0 \\
& \textbf{Ours} & \textbf{86.3} & \textbf{91.8} & \textbf{93.9} & \textbf{95.1} & \textbf{95.7} & \textbf{98.1} & \textbf{98.7} & \textbf{99.2} & \textbf{47.4} & \textbf{52.0} & \textbf{55.5} & 57.2 \\ \bottomrule
\end{tabular}
}
 \label{tab:comparation_ucf}
\end{table*}

\subsection{Comparative Study}
\label{sec:comparative study}
We first reported the results obtained by comparing our method with state-of-the-art methods on EPIC-Kitchens, UCF-HMDB, and Jester datasets in Table~\ref{tab:comparation_Epic}, Table~\ref{table:compare-epic-1020shot} and Table~\ref{tab:comparation_ucf}. 
We observe that multimodal methods often demonstrate superior effectiveness over unimodal methods, due to the complementary nature of multimodal information. 
The proposed MC-LRD is competitive compared to other state-of-the-art unimodal and multimodal methods in 1 and 5-shot settings. 
As shown in Table~\ref{tab:comparation_Epic}, our method demonstrates a more notable improvement on the EPIC-Kitchens dataset, which poses greater challenges due to its fine-grained action data and intricate relationships between modalities. 
The mean accuracies of MC-LRD on the 6 domain adaptation tasks are 49.9\% and 52.2\% in the 1 and 5-shot settings, outperforming the second-best multimodal method by 4.2\% and 5.0\%. 
MC-LRD performs better than the second-best multimodal approach Relamix by 5.3\% in the 1-shot setting on D2$\to$D1 and achieves 4.6\% improvement in the 5-shot setting on D2$\to$D1. 
On the D1$\to$D2 and D3$\to$D1 tasks, the accuracy of MC-LRD is 7.5\% and 6.3\% higher than RelaMix in the 5-shot setting. 
The significant performance improvement of our method on the EPIC-Kitchens dataset demonstrates the effectiveness of disentangling modality-unique and modality-shared components from multimodal data for domain alignment, addressing the challenging fine-grained cross-domain problem. 
As shown in Table~\ref{tab:comparation_ucf}, the proposed MC-LRD achieves improvements of 1.2\% and 1.1\% in the 1-shot setting compared with the second-best multimodal approach (i.e., RelaMix and SSA$^2$lign) on U$\to$H and H$\to$U tasks. 
MC-LRD achieved better performance in the Jester dataset, with improvements of 1.0\% on the 5-shot setting compared to the second-best method. 
Additionally, our model demonstrates 4.7\% and 3.4\% mean accuracy gains compared to RelaMix on the EPIC-Kitchen dataset in both 10-shot and 20-shot settings. 
These results further confirm the effectiveness of our approach. 
These findings indicate that directly applying existing methods does not effectively solve the multimodal Few-Shot Domain Adaptation problem. 
In contrast, MC-LRD achieves better results through learning task-friendly modality-unique and modality-shared components and emphasizes the importance of considering multimodal collaboration. 
Leveraging the collaborative relationships between modalities, our method effectively decomposes and aligns features at diverse domain shift levels, leading to more efficient domain adaptation.

\subsection{Ablation Analysis}
\begin{table*}[t]
\setlength{\tabcolsep}{3pt}
\centering
 \caption{Ablation results on EPIC-Kitchens dataset.}
\label{tab:ablation-epic}
\resizebox{1\linewidth}{!}{
\begin{tabular}{l|cccccccccccc|cc}
\toprule
\multirow{2}{*}{Method} & \multicolumn{2}{c}{D1$\to$D2} & \multicolumn{2}{c}{D1$\to$D3} & \multicolumn{2}{c}{D2$\to$D1} & \multicolumn{2}{c}{D2$\to$D3} & \multicolumn{2}{c}{D3$\to$D1} & \multicolumn{2}{c|}{D3$\to$D2} & \multicolumn{2}{c}{Mean} \\
  & 1-shot & 5-shot & 1-shot & 5-shot & 1-shot & 5-shot & 1-shot & 5-shot & 1-shot & 5-shot & 1-shot & 5-shot & 1-shot & 5-shot \\ \midrule \midrule
w/ Base Model & 45.4  & 48.6  & 39.0  & 42.9  & 40.2  & 43.6  & 43.2  & 47.3  & 44.3  & 46.2  & 50.4  & 53.4  & 43.8 & 47.0 
\\
w/o Decomposer decorrelation loss ($\mathcal{L}_{dd}$) & 48.9 & 51.8 & 41.9 & 44.5 & 46.5 & 48.9 & 50.4 & 51.3 & 45.6 & 49.0 & 53.9 & 56.5 & 47.6 & 50.3 \\
w/o Router decorrelation loss ($\mathcal{L}_{rd}$) & 49.7 & 50.7 & 42.6 & 42.7 & 46.7 & 48.4 & 50.0 & 49.8 & 45.3 & 48.3 & 51.1 & 57.1 & 47.9 & 49.5 \\
w/o Activation consistency loss ($\mathcal{L}_{ac}$) & 50.2 & 50.7 & 42.3 & 43.3 & 46.4 & 48.5 & 48.4 & 51.8 & 45.6 & 46.9 & 55.7 & 56.8 & 48.1 & 49.7 \\
w/o Clip-level decomposers & 47.7  & 52.9  & 43.6  & 39.9  & 44.4  & 48.3  & 49.9  & 49.5  & 42.8  & 45.7  & 52.4  & 55.4 & 46.8 & 48.6 
\\
w/o Video-level decomposers & 51.3  & 53.2  & 42.8  & 43.3  & 46.1  & 48.9  & 50.0  & 52.4  & 46.8  & 48.4  & 53.9  & 56.8  & 48.5 & 50.5 
\\
w/o RGB-unique sub-router  & 48.5  & 52.5  & 39.2  & 43.1  & 46.5  & 48.8  & 50.3  & 50.6  & 46.8  & 49.4  & 55.0  & 54.1  & 47.7 &	49.8 
\\ 
w/o Flow-unique sub-router & 44.6  & 49.7  & 41.5  & 42.3  & 46.3  & 47.1  & 50.4  & 52.6  & 44.9  & 45.9  & 50.3  & 52.5 & 46.3 & 48.3 
\\
w/o Modality-shared sub-router & 47.3  & 52.6  & 40.9  & 41.6 & 46.1  & 48.8  & 50.3  & 51.8  & 44.0  & 45.7  & 52.2  & 52.9  & 46.8 & 48.9 
\\
{\color{black}w/o Progressively shared decomposers} & 51.4 & 53.3 & 44.5 & 45.8 & 46.8 & 49.1 & 50.5 & 52.7 & 48.0 & 52.0 & 55.8 & 57.3 & 49.5 & 51.7 
\\
\textbf{Ours} & \textbf{51.9} & \textbf{53.5} & \textbf{44.9} & \textbf{46.1} & \textbf{46.8} & \textbf{49.4} & \textbf{50.8} & \textbf{53.3} & \textbf{49.1} & \textbf{52.2} & \textbf{55.7} & \textbf{58.7} & \textbf{49.9} & \textbf{52.2} \\ 
\bottomrule 
\end{tabular}
}
\end{table*}

\begin{table*}[!ht]
\setlength{\extrarowheight}{0pt}
\setlength{\tabcolsep}{5pt}
 \caption{Additional ablation results of modality-unique and modality-shared features on Epic-Kitchens dataset.}
\resizebox{\linewidth}{!}{
\begin{tabular}{ccc|cc|cc|cc|cc|cc|cc|cc}
\toprule
\multicolumn{3}{c|}{}  & \multicolumn{2}{c|}{D1$\to$D2} & \multicolumn{2}{c|}{D1$\to$D3} & \multicolumn{2}{c|}{D2$\to$D1} & \multicolumn{2}{c|}{D2$\to$D3} & \multicolumn{2}{c|}{D3$\to$D1} & \multicolumn{2}{c|}{D3$\to$D2} & \multicolumn{2}{c} {Mean} \\
\begin{tabular}[c]{@{}c@{}}RGB-\\  unique\end{tabular} & \begin{tabular}[c]{@{}c@{}}Flow-\\  unique\end{tabular} & \begin{tabular}[c]{@{}c@{}}Modality-\\  shared\end{tabular} & 1-shot  & 5-shot  & 1-shot  & 5-shot  & 1-shot  & 5-shot  & 1-shot  & 5-shot  & 1-shot  & 5-shot  & 1-shot  & 5-shot  & 1-shot  & 5-shot  \\ \midrule \midrule
$\checkmark$  & $\times$  & $\times$  & 36.3  & 37.3  & 38.8  & 39.2  & 35.8  & 32.2  & 44.7  & 45.8  & 31.3  & 35.0  & 42.0  & 45.6 & 38.1 & 39.2 \\
$\times$ & $\checkmark$ & $\times$  & 46.0  & 48.1  & 33.8  & 29.8  & 41.9  & 42.8  & 39.7  & 40.4  & 42.8  & 45.3  & 47.3  & 46.9 & 41.9 & 42.2  \\
$\times$ & $\times$  & $\checkmark$ & 47.3  & 49.0  & 41.1  & 43.8  & 42.2  & 43.7  & 46.4  & 49.6  & 47.9  & 45.3  & 47.4  & 51.9 & 44.4 & 47.2  \\
$\checkmark$ & $\checkmark$  & $\checkmark$  & \textbf{51.9} & \textbf{53.5} & \textbf{44.9} & \textbf{46.1} & \textbf{46.8} & \textbf{49.4} & \textbf{50.8} & \textbf{53.3} & \textbf{49.1} & \textbf{52.2} & \textbf{55.7} & \textbf{58.7} & \textbf{49.9} & \textbf{52.2} \\ \bottomrule
\end{tabular}}
\label{tab:addition-ablation-epic}
\end{table*}

\begin{table}[ht]
 \centering
 \caption{Ablation results on UCF-HMDB and Jester datasets.}
\setlength{\extrarowheight}{0pt}
\setlength{\tabcolsep}{1pt}
 \resizebox{1.0\linewidth}{!}{
\begin{tabular}{l|cc|cc|cc}
\toprule
\multirow{2}{*}{Method} & \multicolumn{2}{c|}{U$\to$H}  & \multicolumn{2}{c|}{H$\to$U}  & \multicolumn{2}{c}{Jester}  \\
 & 1-shot  & 5-shot  & 1-shot  & 5-shot  & 1-shot  & 5-shot  \\ \midrule \midrule
w/ Base Model  & 85.0  & 90.2  & 95.1  & 97.4  & 45.2  & 48.6  \\ 
w/o Decomposer decorrelation Loss  & 84.6 & 89.9 & 93.6 & 98.4 & 45.7 & 50.5 \\
w/o Router decorrelation Loss  & 85.0 & 89.8 & 93.6 & 98.0 & 44.7 & 49.5 \\
w/o Activation consistency Loss  & 84.4 & 88.8 & 94.3 & 97.5 & 45.8 & 49.9  \\ 
w/o clip-level decomposer  & 84.8  & 89.4  & 93.6  & 97.9  & 44.1  & 51.3  \\ 
w/o Video-level decomposer  & 86.1  & 91.6  & 95.4  & {97.9}  & 45.9  & 48.7  \\ 
w/o RGB-unique Router  & 80.6  & 87.0  & 91.4  & 95.5  & 43.4  & 48.2  \\ 
w/o Flow-unique Router  & 83.9  & 86.5  & 92.8  & 96.3  & 44.5  & 46.6  \\ 
w/o modality-shared sub-router  & 82.4  & 89.1  & 92.8  & 96.5  & 45.1  & 48.2  \\ \midrule 
\textbf{Ours} & \textbf{86.3} & \textbf{91.8} & \textbf{95.7} & \textbf{98.1} & \textbf{47.4} & \textbf{52.0} 
\\ \bottomrule
\end{tabular}
}
\label{tab:ablation-ucf-jester}
\end{table}

\begin{table}[t]
\centering
\setlength{\extrarowheight}{0pt}
\setlength{\tabcolsep}{3pt}
 \caption{Additional ablation results of modality-unique and modality-shared features on UCF-HMDB and Jester datasets.}
\resizebox{1\linewidth}{!}{
\begin{tabular}{ccc|cc|cc|cc}
\toprule
\multicolumn{3}{c|}{}  & \multicolumn{2}{c|}{U$\to$H} & \multicolumn{2}{c|}{H$\to$U} & \multicolumn{2}{c}{Jester}  \\
\begin{tabular}[c]{@{}c@{}}RGB-\\  unique\end{tabular} & \begin{tabular}[c]{@{}c@{}}Flow-\\  unique\end{tabular} & \begin{tabular}[c]{@{}c@{}}Modality-\\  shared\end{tabular} & 1-shot  & 5-shot  & 1-shot  & 5-shot  & 1-shot  & 5-shot  \\ \midrule \midrule
$\checkmark$  & $\times$  & $\times$  & 83.3  & 85.9  & 90.3  & 95.2  & 43.2  & 38.7  \\
$\times$ & $\checkmark$ & $\times$  & 72.5  & 85.2  & 84.6  & 87.3  & 41.5  & 48.2  \\
$\times$ & $\times$  & $\checkmark$ & 85.6  & 89.6  & 91.2  & 95.6  & 47.3  & 48.2  \\
$\checkmark$ & $\checkmark$  & $\checkmark$  & \textbf{86.3} & \textbf{91.8} & \textbf{95.7} & \textbf{98.1} & \textbf{47.4} & \textbf{52.0} \\ \bottomrule
\end{tabular}}
 \label{tab:addition-ablation-ucf}
\end{table}

In this section, the effectiveness of the proposed MC-LRD network is further evaluated by analyzing the impact of key components (i.e., multi-scale decomposers, multimodal decomposition routers, and losses $\mathcal{L}_{dd}$,$\mathcal{L}_{rd}$,$\mathcal{L}_{ac}$) on the EPIC-Kitchens dataset. 
The ablation experiments are shown in Table~\ref{tab:ablation-epic} and Table~\ref{tab:ablation-ucf-jester}. 
We first compare the approach utilizing the base model merely. 
It extracts unimodal features from each modality and employs late fusion to obtain final results, trained with both source and few-shot videos. 
It shows the performance drop of 5.3\% on the D3$\to$D2 task in the 1-shot settings, indicating that coupled features are more challenging to align. 
Subsequently, we compare with the variant of MC-LRD that removes the proposed loss terms from the training object in Eq(\ref{eq: objective loss}).
Compared with our complete model, removing the decomposer decorrelation loss $\mathcal{L}_{dd}$, router decorrelation loss $\mathcal{L}_{rd}$ and activation consistency loss $\mathcal{L}_{ac}$, the mean performance drops 2.3\%, 2.0\% and 1.8\% in the 1-shot setting respectively, demonstrating their effectiveness. 
Subsequently, we removed the clip-level decomposer and video-level decomposer respectively, to analyze the individual impacts of them. 
{\color{black}
In addition, we also analyze the impact of the three routers in the MDR module, i.e., RGB-unique router, Flow-unique router, and Moality-shared router. 
These ablation studies reflect the influence of modality-unique or modality-shared features on model performance. 
}
The performance of the variant model declines across most tasks when either the clip-level or video-level decomposer, or the sub-routers in MDR, are absent, underscoring the critical role of these modules. 
We further explore the effectiveness of individual sub-routers in the Multimodal Decomposition Routers and demonstrate the ablation results where every two sub-routers are removed in Table~\ref{tab:addition-ablation-epic} and Table~\ref{tab:addition-ablation-ucf}. 
Note that, when the RGB-unique sub-router and Flow-unique sub-router are removed, the results are based solely on the modality-shared features. 
Conversely, when the modality-shared sub-router is removed, the results are based on the modality-unique features. 
These ablation results underscore the significance of the Multimodal Decomposition Routers in effectively decomposing multimodal features and preserving task-unique information, ultimately enhancing the performance of the model across various tasks and datasets. 
The findings demonstrate that the decomposed features from all three sub-routers contribute to the classification performance and complement each other. 
{\color{black}
We further investigate the impact of the progressively shared decomposers by reconstructing the model without parameter sharing between decomposers in each modality. 
The results demonstrate that the progressively shared design enhances performance. 
}
The above results demonstrate the importance of all components in our method.

\begin{table}[!ht]
\centering
\caption{Comparation results on Epic-Kitchens, UCF-HMDB, and Jester datasets after removing TRX and HyRSM columns. `EF' and `LF' denote the Early-Fusion and Late-Fusion strategies. }
\setlength{\extrarowheight}{0pt}
\setlength{\tabcolsep}{3pt}
\resizebox{1.0\linewidth}{!}{
\begin{tabular}{ll|cc|cc|cc|cc|c}
\toprule
\multicolumn{2}{c|}{Method} & \multicolumn{2}{c|}{TranSVAE~\cite{wei2024unsupervised}} & \multicolumn{2}{c|}{SSA$^2$lign~\cite{xu2023augmenting}} & \multicolumn{2}{c|}{RelaMix~\cite{peng2023relamix}} & \multirow{2}{*}{\textbf{Ours}} \\
\multicolumn{2}{c|}{Fusion strategy} & EF & LF & EF & LF & EF & LF & \\ \midrule\midrule
\multirow{3}{*}{D1-D2} 
& 1-shot & 39.5 & 38.9 & 43.3 & 45.3 & 46.9 & 48.1 & \textbf{51.9} \\
& 5-shot & 37.2 & 44.7 & 43.7 & 47.7 & 46.0 & 46.0 & \textbf{53.5} \\
& 10-shot & 34.1 & 44.0 & 47.1 & 50.5 & 45.6 & 47.5 & \textbf{54.0} \\
& 20-shot & 38.7 & 47.1 & 48.5 & 52.9 & 44.7 & 54.7 & \textbf{56.5} \\ \midrule
\multirow{3}{*}{D1-D3} 
& 1-shot & 41.9 & 34.9 & 41.6 & 39.1 & 38.5 & 39.8 & \textbf{44.9} \\
& 5-shot & 41.7 & 40.0 & 42.4 & 44.0 & 41.7 & 44.1 & \textbf{46.1} \\
& 10-shot & 40.2 & 41.5 & 46.3 & 45.6 & 41.2 & 46.3 & \textbf{48.8} \\
& 20-shot & 37.4 & 38.9 & 45.6 & 46.2 & 40.9 & 46.0 & \textbf{49.5} \\ \midrule
\multirow{3}{*}{D2-D1} 
& 1-shot & 39.8 & 43.0 & 39.8 & 37.2 & 35.4 & 41.5 & \textbf{46.8} \\
& 5-shot & 41.8 & 42.5 & 42.8 & 42.8 & 38.7 & 44.8 & \textbf{49.4} \\
& 10-shot & 40.9 & 42.8 & 43.7 & 44.1 & 39.0 & 46.6 & \textbf{49.6} \\
& 20-shot & 42.5 & 44.8 & 49.2 & 46.2 & 44.5 & 48.0 & \textbf{52.8} \\ \midrule
\multirow{3}{*}{D2-D3} 
& 1-shot & 48.2 & 47.1 & 47.9 & 43.1 & 42.8 & 50.7 & \textbf{50.8} \\
& 5-shot & 45.3 & 50.1 & 51.0 & 45.1 & 45.1 & 46.9 & \textbf{53.3} \\
& 10-shot & 49.3 & 47.6 & 51.6 & 51.3 & 47.5 & 50.6 & \textbf{55.2} \\
& 20-shot & 49.4 & 48.2 & 53.0 & 51.3 & 45.9 & 50.7 & \textbf{56.6} \\ \midrule
\multirow{3}{*}{D3-D1} 
& 1-shot & 36.3 & 36.6 & 44.8 & 39.8 & 39.2 & 42.4 & \textbf{49.1} \\
& 5-shot & 40.5 & 44.8 & 43.4 & 43.7 & 38.7 & 45.9 & \textbf{52.2} \\
& 10-shot & 38.2 & 43.9 & 45.7 & 48.0 & 40.0 & 44.9 & \textbf{52.2} \\
& 20-shot & 39.5 & 43.7 & 40.4 & 50.5 & 40.4 & 50.5 & \textbf{52.7} \\ \midrule
\multirow{3}{*}{D3-D2} 
& 1-shot & 43.9 & 52.0 & 45.9 & 53.1 & 45.3 & 51.6 & \textbf{55.7} \\
& 5-shot & 50.7 & 54.0 & 51.6 & 51.1 & 48.2 & 55.6 & \textbf{58.7} \\
& 10-shot & 47.3 & 53.9 & 51.1 & 55.1 & 44.6 & 54.1 & \textbf{58.3} \\
& 20-shot & 53.6 & 54.7 & 52.8 & 52.4 & 46.3 & 56.0 & \textbf{58.4} \\ \midrule
\multirow{3}{*}{Mean} 
& 1-shot & 41.6 & 42.1 & 43.9 & 42.9 & 41.4 & 45.7 & \textbf{49.9} \\
& 5-shot & 42.9 & 46.0 & 45.8 & 45.7 & 43.1 & 47.2 & \textbf{52.2} \\
& 10-shot & 41.7 & 45.6 & 47.6 & 49.1 & 43.0 & 48.3 & \textbf{53.0} \\
& 20-shot & 45.2 & 47.0 & 50.3 & 50.1 & 44.7 & 51.0 & \textbf{55.0} \\ \midrule
\multirow{3}{*}{U-H} 
& 1-shot & 75.0 & 77.5 & 84.4 & 85.3 & 82.0 & 85.1 & \textbf{86.3} \\
& 5-shot & 78.1 & 85.6 & 91.7 & 91.7 & 87.6 & 90.3 & \textbf{92.8} \\
& 10-shot & 76.9 & 81.1 & 93.1 & 93.3 & 90.5 & 91.1 & \textbf{93.9} \\
& 20-shot & 79.7 & 81.9 & 95.0 & 94.2 & 90.4 & 94.0 & \textbf{95.1} \\ \midrule
\multirow{3}{*}{H-U} 
& 1-shot & 72.2 & 69.0 & 93.7 & 93.2 & 92.4 & 94.6 & \textbf{95.7} \\
& 5-shot & 81.6 & 84.8 & 96.5 & 97.2 & 97.0 & 97.4 & \textbf{98.1} \\
& 10-shot & 82.7 & 81.1 & 98.1 & 99.3 & 96.1 & 98.3 & \textbf{98.7} \\
& 20-shot & 81.3 & 86.2 & 99.3 & 99.5 & 96.6 & 98.4 & \textbf{99.2} \\ \midrule
\multirow{3}{*}{J-J} 
& 1-shot & 38.1 & 42.9 & 44.5 & 45.9 & 42.5 & 47.0 & \textbf{47.4} \\
& 5-shot & 39.0 & 41.7 & 44.9 & 51.0 & 45.9 & 51.0 & \textbf{52.0} \\
& 10-shot & 38.4 & 43.1 & 50.7 & 53.6 & 47.0 & 55.4 & \textbf{55.5} \\
& 20-shot & 38.9 & 42.1 & 55.4 & 55.8 & 49.2 & \textbf{58.0} & {57.2} \\ \bottomrule
\end{tabular}}
\label{tab:comparation_EF_updated}
\end{table}
\subsection{Further Remarks}
\label{sec:experiment-further}

\begin{figure}[ht]
    \centering
    \includegraphics[width=0.9\linewidth]{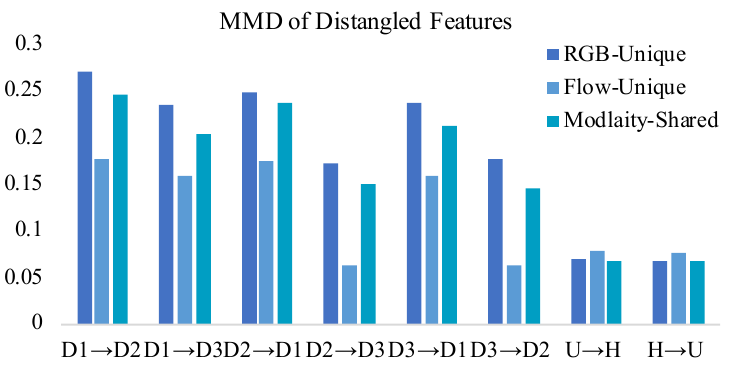}
    \caption{\color{black}
    Quantitative analysis of domain shifts in modality-unique and modality-shared features measured on the EPIC-Kitchens and UCF-HMDB datasets}
    \label{fig:decomposed_mmd}
\end{figure}

{
\color{black}
\subsubsection{
Quantitative Results of Domain Shifts of Modality-unique and Modality-shared Features}
We demonstrate the quantization result of the domain shift observed in each decomposed feature of video data in Figure~\ref{fig:decomposed_mmd}. 
To illustrate the inherent domain shift, we exclude the alignment loss $\mathcal{L}_{ada}$ from the overall objective in Eq(\ref{eq: objective loss}) and train the MC-LRD model solely on the source training set to ensure that domain alignment is not yet been applied. 
Specifically, we report the Maximum Mean Discrepancy (MMD) between the feature representations from the test sets of the source and target domains on the EPIC-Kitchens and UCF-HMDB datasets. 
For the EPIC-Kitchens dataset, RGB-unique features exhibit greater domain shifts compared to Flow-unique features, while the domain shifts in modality-shared features consistently lie between those of modality-unique features for each modality. 
This observation suggests that RGB modality in EPIC-Kitchens contains greater variability and noise. 
Although videos across domains reflect similar actions, variations in the diverse kitchen environments contribute to a more obvious domain shift in RGB-unique features, whereas motion features tend to be more consistent. 
The modality-shared features exclude modality-specific noise inherent to the RGB data, but they remain influenced by variations in actor behavior and thus exhibit a domain shift that falls between the shifts observed in modality-unique features. 
In contrast, in the UCF-HMDB dataset, the domain shift in modality-shared features is relatively smaller than that in modality-unique features.
UCF-HMDB primarily comprises online videos that typically contain less consistent motion information, as these videos originate from diverse, uncurated sources rather than from systematically collected and annotated datasets (such as EPIC-Kitchens). 
The appearance features exhibit greater inter-domain invariance compared to frequently varying motion features. 
Modality-shared features capture semantically rich dependencies and exhibit greater stability cross domains.
}

\subsubsection{Comparison Results with Early-Fusion Multimodal Baselines}

To provide a comprehensive evaluation, we construct multimodal baselines employing both early-fusion and late-fusion strategies. 
The early-fusion strategy combines multimodal features after feature extraction, aiming to integrate information from all modalities before domain adaptation.
In contrast, the late-fusion approach merges features following domain adaptation for each individual modality, enabling each modality to contribute independently to the final representation. 
In section~\ref{sec:comparative study}, we report results from late-fusion baselines, as these methods demonstrate superior performance across the primary experimental settings compared to early-fusion baselines. 
This section provides an in-depth comparison between our approach and both early-fusion and late-fusion multimodal baselines. Table~\ref{tab:comparation_EF_updated} presents these results, offering further insights into the effectiveness of our method relative to standard multimodal fusion strategies.

\begin{table*}[h]
 \centering
 \caption{Comparison results of multimodal feature decomposition strategies. }
 \resizebox{1.0\linewidth}{!}{
\begin{tabular}{l|cccccccccccc|cc}
\toprule
\multirow{2}{*}{Decomposition Strategy} & \multicolumn{2}{c}{D1$\to$D2} & \multicolumn{2}{c}{D1$\to$D3} & \multicolumn{2}{c}{D2$\to$D1} & \multicolumn{2}{c}{D2$\to$D3} & \multicolumn{2}{c}{D3$\to$D1} & \multicolumn{2}{c|}{D3$\to$D2} & \multicolumn{2}{c}{Mean}  \\
  & 1-shot  & 5-shot  & 1-shot  & 5-shot  & 1-shot  & 5-shot  & 1-shot  & 5-shot  & 1-shot  & 5-shot  & 1-shot  & 5-shot  & 1-shot  & 5-shot  \\ \midrule
MLP\cite{gu20203d, yao2024drfuse} & 46.5  & 49.7  & 40.7  & 42.8  & 39.4  & 43.1  & 46.6  & 47.7  & 43.0  & 45.0  & 49.8  & 52.2  & 44.3  & 47.9  \\
Transformer\cite{gu20203d, yao2024drfuse} & 47.2  & 50.5  & 41.6  & 44.0  & 45.8  & 46.9  & 50.4  & 51.4  & 45.9  & 47.4  & 55.7  & 56.7  & 45.7  & 48.0  \\
\textbf{Ours} & \textbf{51.9} & \textbf{53.5} & \textbf{44.9} & \textbf{46.1} & \textbf{46.8} & \textbf{49.4} & \textbf{50.8} & \textbf{53.3} & \textbf{49.1} & \textbf{52.2} & \textbf{55.7} & \textbf{58.7} & \textbf{49.9} & \textbf{52.2}  
\\ \bottomrule
\end{tabular}}
 \label{tab:disentangle}
\end{table*}

\subsubsection{Analysis of Multimodal Decomposition Strategy.}
To further investigate the effectiveness of the proposed MC-LRD framework, we conduct a comparative analysis with existing feature decomposition strategies in the 1-shot setting on the EPIC-Kitchens dataset. 
Specifically, we compare our MC-LRD network to conventional disentanglement approaches similar to those in ~\cite{gu20203d, yao2024drfuse}. 
In these baseline methods, inputs from each modality are independently encoded into modality-unique and modality-shared features using separate models, which are implemented as multilayer perceptrons (MLPs) and Transformer networks. 
This independent encoding approach seeks to disentangle information within each modality separately, before attempting cross-modal alignment.
We ensure fair comparison by training all models using the similar objective function as outlined for our model in Eq(\ref{eq: objective loss}), including classification loss, adversarial domain alignment loss, and disentanglement loss. This shared training objective allows us to directly assess the impact of the decomposition strategy on model performance.
As presented in Table~\ref{tab:disentangle}, our proposed multimodal decomposition approach achieves significantly higher performance than conventional decomposition methods, particularly in capturing both modality-specific and shared features under domain adaptation conditions. 
These results demonstrate the advantage of our approach in effectively aligning cross-modal information and underscore the benefits of our multimodal decomposition strategy for few-shot action recognition tasks.

\begin{figure}
 \begin{center}
\includegraphics[width=\linewidth]{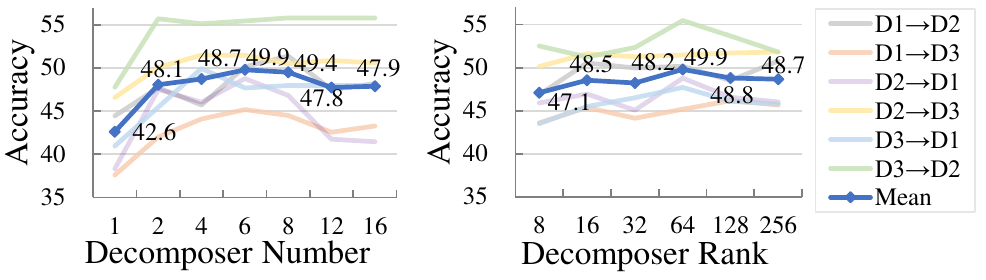}
 \end{center}
 \caption{
Analysis of the number and the rank of decomposers on the EPIC-Kitchens Dataset. }
 \label{fig:experiment_hyperparameters}
\end{figure}

\subsubsection{Sensitivity to Hyperparameters.}
We analyze the decomposer rank ($d_{ra}$) and the decomposer number ($N_c$ and $N_v$) in the MC-LRD module on the EPIC-Kitchens dataset in the 1-shot setting. 
We denote $N_c=N_v$ as $N$. 
Figure~\ref{fig:experiment_hyperparameters} shows the hyperparameter sensitivity analysis. 
For the decomposer rank ($d_{ra}$), optimal performance is achieved at $d_{ra}=64$ when varying from 8 to 256. 
Regarding the number of decomposers ($N$), we explore a range from 1 to 16, noting that decoupling is not performed when $N=1$. 
Insufficient decomposers are inadequate to learn complex decomposed features, while excessive decomposers can increase model complexity and overfitting risk. 
We compromise by selecting $d_{ra}=64$ and $N=6$ for our model, considering both performance and computational efficiency.

\begin{figure}
 \begin{center}
\includegraphics[width=\linewidth]{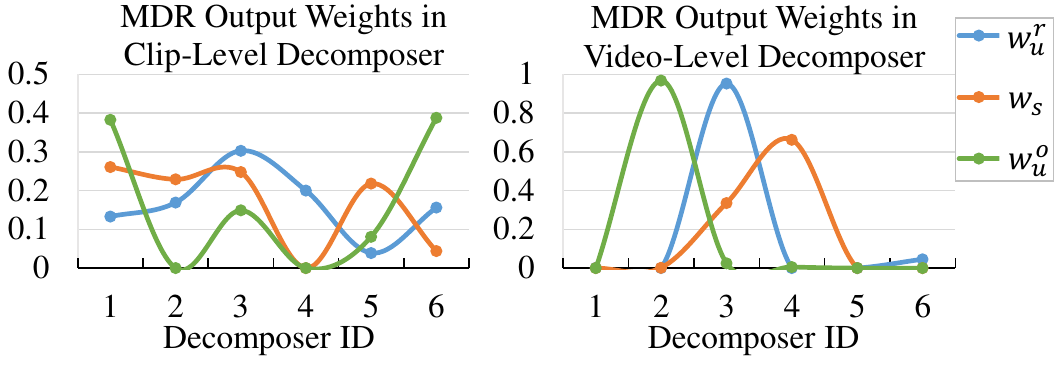}
 \end{center}
 \caption{Visualization of weights output from MDR. }
 \label{fig:weights}
 
\end{figure}
\begin{figure}[!t]
 \begin{center}
\includegraphics[width=\linewidth]{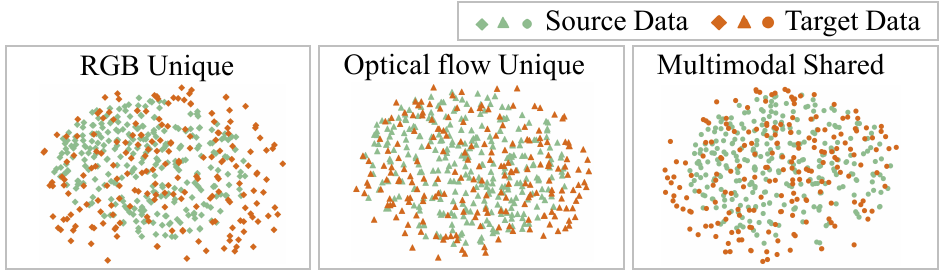}
 \end{center}
 \caption{Visualization of the distribution modality-unique and modality-shared features extracted by MC-LRD. }
 \label{fig: visualization}
\end{figure}

\subsubsection{Qualitative Results.}
To visually demonstrate the effectiveness of the proposed MC-LRD, we depict the mean weights output from the first layer MDR in Figure~\ref{fig:weights} on the D1$\to$D2 task in the test set of the EPIC-Kitchens dataset. 
These weights ($w_u^r$, $w_s$, $w_u^o$) indicate the activation preferences of sub-routers. 
As illustrated, in both clip-level and video-level decomposers, there is a notable difference in the weights assigned to decomposers for modality-unique and modality-shared components, indicating effective decomposition achieved by our model. 
We additionally showcase the distribution of multimodal cross-domain data features optimized by our method. 
In comparison to Figure 1 in the main text, our approach demonstrates notably effective alignment for both unimodal and multimodal features. 
As illustrated in Figure~\ref{fig: visualization}, we showcase the distribution of data in the source and target domain test sets. 
decomposed modality-unique and modality-shared features demonstrate alignment across domains.

\subsubsection{Complexity Analysis}

{\color{black}
We compare the training costs of our model (Basel\#M+Adapt) with unimodal (Basel\#U) and multimodal (Basel\#M) baselines in Figure 8. 
We implement the experiments with the same batch size of 128 for 50 epochs on the $D1\to D2$ task. 
We establish two baseline methods: Basel\#U and Basel\#M. 
Both leverage the base model architecture, aligning with the ``Base model" entries in Table IV of our ablation study. 
Basel\#U functions as the unimodal baseline, processing exclusively RGB inputs. 
Basel\#M operates as the multimodal baseline, utilizing late fusion to integrate cross-modal features. 
For multi-modal methods (Basel\#M and Basel\#M+Adapt), our adaptation approach significantly reduces the memory usage, while maintaining an acceptable time cost, because our method only updates the parameters of the decomposer during training, instead of all network parameters.
In contrast to the single-modal method (Basel\#U+Adapt), the incorporation of the multi-modal input (i.e., optical flow) increases the acceptable memory usage, yet yields significant performance benefits (see Section IV.B). 
Multimodal methods did not increase the training time obviously, as features of RGB and optical flow are decomposed in parallel. 
}

\begin{figure}[t]
    \centering
    \includegraphics[width=0.8\linewidth]{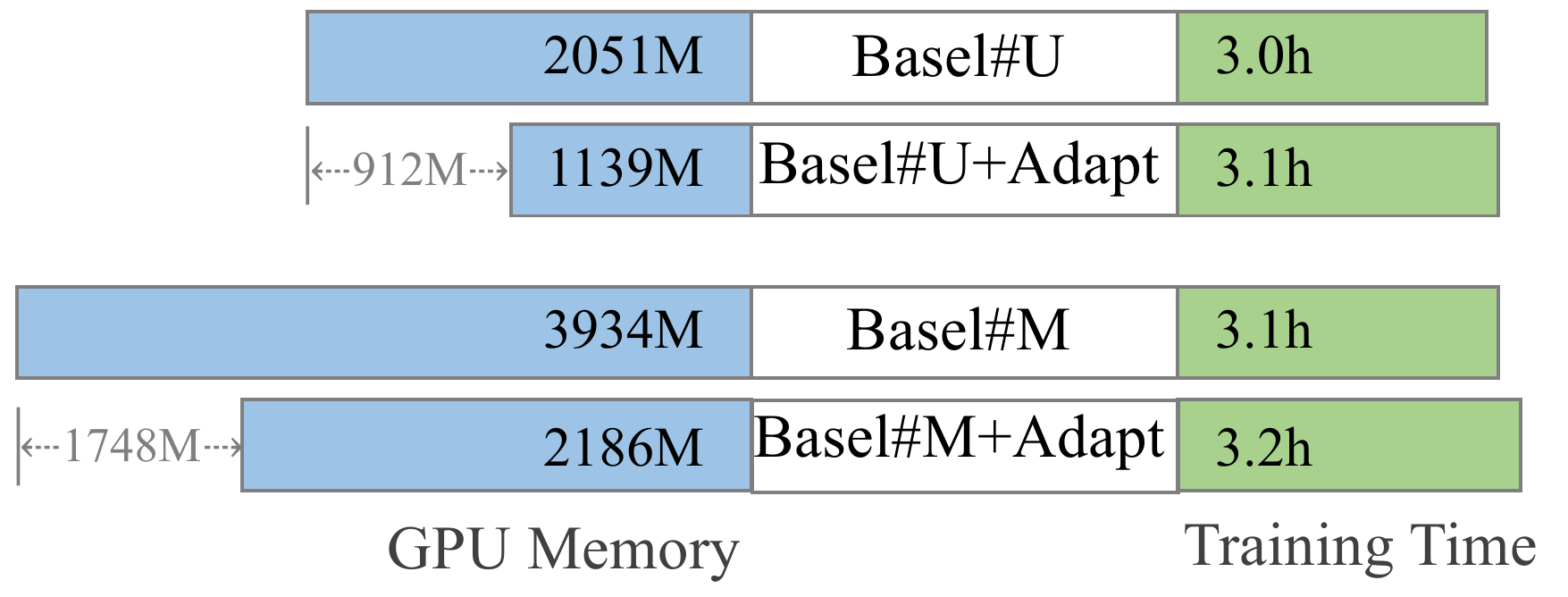}
    \caption{Training cost of MC-LRD compared to baselines. }
    \label{fig:training cost}
\end{figure}

We additionally compare our method with multimodal baselines in terms of the number of trainable parameters, multiply-accumulate operations (MACs), and the average inference time per video in Table~\ref{tab:complexity}. 
\begin{table}[!t]
\centering
\caption{Comparison results on model complexity. MC-LRD-P and MC-LRD-A represent the MC-LRD in pre-training and adaptation steps respectively. }
\resizebox{1.0\linewidth}{!}{
\begin{tabular}{l|ccc}
\toprule
\textbf{Methods}  & \textbf{Params (M)} & \textbf{MACs (G)} & \textbf{Inference Time (s)} \\ \midrule \midrule
TRX~\cite{perrett2021trx}  & 16.78  & 70.93 & 0.2071  \\
HyRSM~\cite{wang2022hyrsm} & 12.09  & 7.08  & 0.0146  \\
TranSVA~\cite{wei2024unsupervised} & 78.11  & 18.75 & 0.0510  \\
SSA$^2$lign~\cite{xu2023augmenting}  & 67.24  & 25.83 & 0.6376  \\
RelaMix~\cite{peng2023relamix}  & 119.64  & 93.86 & 2.3404  \\\midrule
MC-LRD-P & 9.48 & 3.23  & 0.0106  \\ 
MC-LRD-A & 3.01 & 4.91  & 0.0375  \\
\bottomrule
\end{tabular}}
\label{tab:complexity}
%
\end{table}
We observe that our method significantly reduces the required number of training parameters, especially in the main adaptation step. 
Leveraging existing methods for the FSVDA task directly necessitates executing multimodal models in parallel, inevitably leading to an escalation in trainable parameters and inference time. 
In contrast, MC-LRD adopts LoRA decomposers combined with a two-stage training strategy, resulting in a significant reduction in parameter count and computational complexity. 
Especially, in the main adaptation phase (MC-LRD-A), the trainable parameters are only 3.01M, substantially lower compared to the most competitive baseline (i.e., HyRSM with 12.09M). 
In terms of MACs and inference speed, the MC-LRD achieves an acceptable and competitive performance. 
Despite the slower inference efficiency compared to HyRSM, MC-LRD exhibits a significant advantage in performance on the FSVDA task. 
These results further highlight the effectiveness of our method in achieving superior inference performance with limited complexity.

\section{Conclusion}

In this paper, we address the challenge of learning both modality-unique and modality-shared features from multimodal video data, aiming to improve the effectiveness of Few-Shot Video Domain Adaptation (FSVDA). 
To this end, we propose a novel framework, Modality-Collaborative Low-Rank Decomposers (MC-LRD), designed to enable more efficient and adaptive feature decomposition across modalities. 
The MC-LRD framework introduces modality-dependent decomposers that are selectively activated by a multimodal decomposition router, ensuring that modality-unique and modality-shared features are effectively disentangled and independently optimized. 
This selective decomposition is further supported by orthogonal decorrelation losses, which help to preserve feature independence across modalities, and a cross-domain activation consistency loss to facilitate robust domain alignment in cross-modal settings.
Extensive experiments on three public datasets validate the effectiveness of the proposed method, demonstrating its superior performance in capturing and aligning multimodal features under challenging few-shot scenarios. 
Looking forward, we plan to extend our network to scenarios where certain modalities may be missing, further enhancing the robustness and real-world applicability of our approach across diverse multimodal contexts.
{
\color{black}
Our method considers the RGB and optical flow modalities as input modalities, being restricted to two-modal scenarios while failing to generalize to multimodal scenarios with more than two modalities. 
Future work could explore the inclusion of other modalities, such as audio and depth information, to address this challenging task. 
Additionally, adapting the algorithm to tackle more challenging tasks, such as video action localization and video behavior prediction, is also a promising and practical research direction for the future. 
}

\bibliographystyle{plain}
\bibliography{collection}

@inproceedings{munro2020multi,
  title={Multi-modal domain adaptation for fine-grained action recognition},
  author={Munro, Jonathan and Damen, Dima},
  booktitle={Proceedings of the IEEE/CVF conference on computer vision and pattern recognition},
  pages={122--132},
  year={2020}
}

@inproceedings{song2021spatio,
  title={Spatio-temporal contrastive domain adaptation for action recognition},
  author={Song, Xiaolin and Zhao, Sicheng and Yang, Jingyu and Yue, Huanjing and Xu, Pengfei and Hu, Runbo and Chai, Hua},
  booktitle={Proceedings of the IEEE/CVF conference on computer vision and pattern recognition},
  pages={9787--9795},
  year={2021}
}

@inproceedings{kim2021learning,
  title={Learning cross-modal contrastive features for video domain adaptation},
  author={Kim, Donghyun and Tsai, Yi-Hsuan and Zhuang, Bingbing and Yu, Xiang and Sclaroff, Stan and Saenko, Kate and Chandraker, Manmohan},
  booktitle={Proceedings of the IEEE/CVF International Conference on Computer Vision},
  pages={13618--13627},
  year={2021}
}

@inproceedings{lv2021differentiated,
  title={Differentiated learning for multi-modal domain adaptation},
  author={Lv, Jianming and Liu, Kaijie and He, Shengfeng},
  booktitle={Proceedings of the 29th ACM international conference on multimedia},
  pages={1322--1330},
  year={2021}
}

@inproceedings{da2022dual,
  title={Dual-head contrastive domain adaptation for video action recognition},
  author={Da Costa, Victor G Turrisi and Zara, Giacomo and Rota, Paolo and Oliveira-Santos, Thiago and Sebe, Nicu and Murino, Vittorio and Ricci, Elisa},
  booktitle={Proceedings of the IEEE/CVF Winter Conference on Applications of Computer Vision},
  pages={1181--1190},
  year={2022}
}

@inproceedings{huang2022relative,
  title={Relative alignment network for source-free multimodal video domain adaptation},
  author={Huang, Yi and Yang, Xiaoshan and Zhang, Ji and Xu, Changsheng},
  booktitle={Proceedings of the 30th ACM International Conference on Multimedia},
  pages={1652--1660},
  year={2022}
}

@inproceedings{yang2022interact,
  title={Interact before align: Leveraging cross-modal knowledge for domain adaptive action recognition},
  author={Yang, Lijin and Huang, Yifei and Sugano, Yusuke and Sato, Yoichi},
  booktitle={Proceedings of the IEEE/CVF conference on computer vision and pattern recognition},
  pages={14722--14732},
  year={2022}
}

@inproceedings{zhang2022audio,
  title={Audio-adaptive activity recognition across video domains},
  author={Zhang, Yunhua and Doughty, Hazel and Shao, Ling and Snoek, Cees GM},
  booktitle={Proceedings of the IEEE/CVF Conference on Computer Vision and Pattern Recognition},
  pages={13791--13800},
  year={2022}
}

@inproceedings{li2023source,
  title={Source-free video domain adaptation with spatial-temporal-historical consistency learning},
  author={Li, Kai and Patel, Deep and Kruus, Erik and Min, Martin Renqiang},
  booktitle={Proceedings of the IEEE/CVF Conference on Computer Vision and Pattern Recognition},
  pages={14643--14652},
  year={2023}
}

@inproceedings{zara2023unreasonable,
  title={The unreasonable effectiveness of large language-vision models for source-free video domain adaptation},
  author={Zara, Giacomo and Conti, Alessandro and Roy, Subhankar and Lathuili{\`e}re, St{\'e}phane and Rota, Paolo and Ricci, Elisa},
  booktitle={Proceedings of the IEEE/CVF International Conference on Computer Vision},
  pages={10307--10317},
  year={2023}
}

@inproceedings{yin2022mix,
  title={Mix-dann and dynamic-modal-distillation for video domain adaptation},
  author={Yin, Yuehao and Zhu, Bin and Chen, Jingjing and Cheng, Lechao and Jiang, Yu-Gang},
  booktitle={Proceedings of the 30th ACM International Conference on Multimedia},
  pages={3224--3233},
  year={2022}
}

@inproceedings{lin2022cycda,
  title={Cycda: Unsupervised cycle domain adaptation to learn from image to video},
  author={Lin, Wei and Kukleva, Anna and Sun, Kunyang and Possegger, Horst and Kuehne, Hilde and Bischof, Horst},
  booktitle={European Conference on Computer Vision},
  pages={698--715},
  year={2022},
  organization={Springer}
}

@inproceedings{choi2020shuffle,
  title={Shuffle and attend: Video domain adaptation},
  author={Choi, Jinwoo and Sharma, Gaurav and Schulter, Samuel and Huang, Jia-Bin},
  booktitle={Computer Vision--ECCV 2020: 16th European Conference, Glasgow, UK, August 23--28, 2020, Proceedings, Part XII 16},
  pages={678--695},
  year={2020},
  organization={Springer}
}

@inproceedings{dasgupta2022overcoming,
  title={Overcoming label noise for source-free unsupervised video domain adaptation},
  author={Dasgupta, Avijit and Jawahar, CV and Alahari, Karteek},
  booktitle={Proceedings of the Thirteenth Indian Conference on Computer Vision, Graphics and Image Processing},
  pages={1--9},
  year={2022}
}

@article{wei2024unsupervised,
  title={Unsupervised Video Domain Adaptation for Action Recognition: A Disentanglement Perspective},
  author={Wei, Pengfei and Kong, Lingdong and Qu, Xinghua and Ren, Yi and Xu, Zhiqiang and Jiang, Jing and Yin, Xiang},
  journal={Advances in Neural Information Processing Systems},
  volume={36},
  year={2024}
}

@inproceedings{kang2019contrastive,
  title={Contrastive adaptation network for unsupervised domain adaptation},
  author={Kang, Guoliang and Jiang, Lu and Yang, Yi and Hauptmann, Alexander G},
  booktitle={Proceedings of the IEEE/CVF conference on computer vision and pattern recognition},
  pages={4893--4902},
  year={2019}
}

@inproceedings{xu2019d,
  title={d-sne: Domain adaptation using stochastic neighborhood embedding},
  author={Xu, Xiang and Zhou, Xiong and Venkatesan, Ragav and Swaminathan, Gurumurthy and Majumder, Orchid},
  booktitle={Proceedings of the IEEE/CVF Conference on Computer Vision and Pattern Recognition},
  pages={2497--2506},
  year={2019}
}

@article{motiian2017few,
  title={Few-shot adversarial domain adaptation},
  author={Motiian, Saeid and Jones, Quinn and Iranmanesh, Seyed and Doretto, Gianfranco},
  journal={Advances in neural information processing systems},
  volume={30},
  year={2017}
}

@inproceedings{xu2023augmenting,
  title={Augmenting and Aligning Snippets for Few-Shot Video Domain Adaptation},
  author={Xu, Yuecong and Yang, Jianfei and Zhou, Yunjiao and Chen, Zhenghua and Wu, Min and Li, Xiaoli},
  booktitle={Proceedings of the IEEE/CVF International Conference on Computer Vision},
  pages={13445--13456},
  year={2023}
}

@article{li2021supervised,
  title={Supervised domain adaptation for few-shot radar-based human activity recognition},
  author={Li, Xinyu and He, Yuan and Zhang, J Andrew and Jing, Xiaojun},
  journal={IEEE Sensors Journal},
  volume={21},
  number={22},
  pages={25880--25890},
  year={2021},
  publisher={IEEE}
}

@article{gao2020pairwise1,
  title={A pairwise attentive adversarial spatiotemporal network for cross-domain few-shot action recognition-R2},
  author={Gao, Zan and Guo, Leming and Guan, Weili and Liu, An-An and Ren, Tongwei and Chen, Shengyong},
  journal={IEEE Transactions on Image Processing},
  volume={30},
  pages={767--782},
  year={2020},
  publisher={IEEE}
}

@article{gao2020pairwise2,
  title={Pairwise two-stream convnets for cross-domain action recognition with small data},
  author={Gao, Zan and Guo, Leming and Ren, Tongwei and Liu, An-An and Cheng, Zhi-Yong and Chen, Shengyong},
  journal={IEEE Transactions on Neural Networks and Learning Systems},
  volume={33},
  number={3},
  pages={1147--1161},
  year={2020},
  publisher={IEEE}
}

@inproceedings{peng2023relamix,
  title={Exploring Few-Shot Adaptation for Activity Recognition on Diverse Domains},
  author={Kunyu Peng and Di Wen and David Schneider and Jiaming Zhang and Kailun Yang and M. Saquib Sarfraz and Rainer Stiefelhagen and Alina Roitberg},
  publisher={ArXiv},
  year={2023},
  url={https://doi.org/10.48550/arXiv.2305.08420}
}

@article{wang2023cross,
  title={Cross-domain few-shot action recognition with unlabeled videos},
  author={Wang, Xiang and Zhang, Shiwei and Qing, Zhiwu and Lv, Yiliang and Gao, Changxin and Sang, Nong},
  journal={Computer Vision and Image Understanding},
  volume={233},
  pages={103737},
  year={2023},
  publisher={Academic Press}
}

@inproceedings{samarasinghe2023cdfsl,
  title={Cdfsl-v: Cross-domain few-shot learning for videos},
  author={Samarasinghe, Sarinda and Rizve, Mamshad Nayeem and Kardan, Navid and Shah, Mubarak},
  booktitle={Proceedings of the IEEE/CVF International Conference on Computer Vision},
  pages={11643--11652},
  year={2023}
}

@inproceedings{li2022cross,
  title={Cross-domain few-shot learning with task-specific adapters},
  author={Li, Wei-Hong and Liu, Xialei and Bilen, Hakan},
  booktitle={Proceedings of the IEEE/CVF conference on computer vision and pattern recognition},
  pages={7161--7170},
  year={2022}
}

@inproceedings{zhao2021domain,
  title={Domain-adaptive few-shot learning},
  author={Zhao, An and Ding, Mingyu and Lu, Zhiwu and Xiang, Tao and Niu, Yulei and Guan, Jiechao and Wen, Ji-Rong},
  booktitle={Proceedings of the IEEE/CVF Winter Conference on Applications of Computer Vision},
  pages={1390--1399},
  year={2021}
}

@inproceedings{gao2022acrofod,
  title={Acrofod: An adaptive method for cross-domain few-shot object detection},
  author={Gao, Yipeng and Yang, Lingxiao and Huang, Yunmu and Xie, Song and Li, Shiyong and Zheng, Wei-Shi},
  booktitle={European Conference on Computer Vision},
  pages={673--690},
  year={2022},
  organization={Springer}
}

@article{shazeer2017outrageously,
  title={Outrageously large neural networks: The sparsely-gated mixture-of-experts layer},
  author={Shazeer, Noam and Mirhoseini, Azalia and Maziarz, Krzysztof and Davis, Andy and Le, Quoc and Hinton, Geoffrey and Dean, Jeff},
  journal={arXiv preprint arXiv:1701.06538},
  year={2017}
}

@article{fedus2022switch,
  title={Switch transformers: Scaling to trillion parameter models with simple and efficient sparsity},
  author={Fedus, William and Zoph, Barret and Shazeer, Noam},
  journal={Journal of Machine Learning Research},
  volume={23},
  number={120},
  pages={1--39},
  year={2022}
}

@article{jacobs1991adaptive,
  title={Adaptive mixtures of local experts},
  author={Jacobs, Robert A and Jordan, Michael I and Nowlan, Steven J and Hinton, Geoffrey E},
  journal={Neural computation},
  volume={3},
  number={1},
  pages={79--87},
  year={1991},
  publisher={MIT Press}
}

@article{jordan1994hierarchical,
  title={Hierarchical mixtures of experts and the EM algorithm},
  author={Jordan, Michael I and Jacobs, Robert A},
  journal={Neural computation},
  volume={6},
  number={2},
  pages={181--214},
  year={1994},
  publisher={MIT Press}
}

@inproceedings{du2022glam,
  title={Glam: Efficient scaling of language models with mixture-of-experts},
  author={Du, Nan and Huang, Yanping and Dai, Andrew M and Tong, Simon and Lepikhin, Dmitry and Xu, Yuanzhong and Krikun, Maxim and Zhou, Yanqi and Yu, Adams Wei and Firat, Orhan and others},
  booktitle={International Conference on Machine Learning},
  pages={5547--5569},
  year={2022},
  organization={PMLR}
}

@article{riquelme2021scaling,
  title={Scaling vision with sparse mixture of experts},
  author={Riquelme, Carlos and Puigcerver, Joan and Mustafa, Basil and Neumann, Maxim and Jenatton, Rodolphe and Susano Pinto, Andr{\'e} and Keysers, Daniel and Houlsby, Neil},
  journal={Advances in Neural Information Processing Systems},
  volume={34},
  pages={8583--8595},
  year={2021}
}

@inproceedings{wu2023mole,
  title={MoLE: Mixture of LoRA Experts},
  author={Wu, Xun and Huang, Shaohan and Wei, Furu},
  booktitle={The Twelfth International Conference on Learning Representations},
  year={2023}
}

@article{chen2024llava,
  title={Llava-mole: Sparse mixture of lora experts for mitigating data conflicts in instruction finetuning mllms},
  author={Chen, Shaoxiang and Jie, Zequn and Ma, Lin},
  journal={arXiv preprint arXiv:2401.16160},
  year={2024}
}

@inproceedings{hu2021lora,
  title={LoRA: Low-Rank Adaptation of Large Language Models},
  author={Hu, Edward J and Wallis, Phillip and Allen-Zhu, Zeyuan and Li, Yuanzhi and Wang, Shean and Wang, Lu and Chen, Weizhu and others},
  booktitle={International Conference on Learning Representations},
  year={2021}
}

@inproceedings{kuehne2011hmdb,
  title={HMDB: a large video database for human motion recognition},
  author={Kuehne, Hildegard and Jhuang, Hueihan and Garrote, Est{\'\i}baliz and Poggio, Tomaso and Serre, Thomas},
  booktitle={2011 International conference on computer vision},
  pages={2556--2563},
  year={2011},
  organization={IEEE}
}

@article{soomro2012ucf,
  title={UCF101: A dataset of 101 human actions classes from videos in the wild},
  author={Soomro, Khurram and Zamir, Amir Roshan and Shah, Mubarak},
  journal={arXiv preprint arXiv:1212.0402},
  year={2012}
}

@inproceedings{damen2018scaling,
  title={Scaling egocentric vision: The epic-kitchens dataset},
  author={Damen, Dima and Doughty, Hazel and Farinella, Giovanni Maria and Fidler, Sanja and Furnari, Antonino and Kazakos, Evangelos and Moltisanti, Davide and Munro, Jonathan and Perrett, Toby and Price, Will and others},
  booktitle={Proceedings of the European conference on computer vision (ECCV)},
  pages={720--736},
  year={2018}
}

@inproceedings{materzynska2019jester,
  title={The jester dataset: A large-scale video dataset of human gestures},
  author={Materzynska, Joanna and Berger, Guillaume and Bax, Ingo and Memisevic, Roland},
  booktitle={Proceedings of the IEEE/CVF international conference on computer vision workshops},
  pages={0--0},
  year={2019}
}

@inproceedings{pan2020adversarial,
  title={Adversarial cross-domain action recognition with co-attention},
  author={Pan, Boxiao and Cao, Zhangjie and Adeli, Ehsan and Niebles, Juan Carlos},
  booktitle={Proceedings of the AAAI conference on artificial intelligence},
  volume={34},
  number={07},
  pages={11815--11822},
  year={2020}
}

@inproceedings{carreira2017quo,
  title={Quo vadis, action recognition? a new model and the kinetics dataset},
  author={Carreira, Joao and Zisserman, Andrew},
  booktitle={proceedings of the IEEE Conference on Computer Vision and Pattern Recognition},
  pages={6299--6308},
  year={2017}
}

@article{kay2017kinetics,
  title={The kinetics human action video dataset},
  author={Kay, Will and Carreira, Joao and Simonyan, Karen and Zhang, Brian and Hillier, Chloe and Vijayanarasimhan, Sudheendra and Viola, Fabio and Green, Tim and Back, Trevor and Natsev, Paul and others},
  journal={arXiv preprint arXiv:1705.06950},
  year={2017}
}

@article{gonzalez2018image,
  title={Image-to-image translation for cross-domain disentanglement},
  author={Gonzalez-Garcia, Abel and Van De Weijer, Joost and Bengio, Yoshua},
  journal={Advances in neural information processing systems},
  volume={31},
  year={2018}
}

@inproceedings{liu2021smoothing,
  title={Smoothing the disentangled latent style space for unsupervised image-to-image translation},
  author={Liu, Yahui and Sangineto, Enver and Chen, Yajing and Bao, Linchao and Zhang, Haoxian and Sebe, Nicu and Lepri, Bruno and Wang, Wei and De Nadai, Marco},
  booktitle={Proceedings of the IEEE/CVF Conference on Computer Vision and Pattern Recognition},
  pages={10785--10794},
  year={2021}
}

@article{cheng2020improving,
  title={Improving disentangled text representation learning with information-theoretic guidance},
  author={Cheng, Pengyu and Min, Martin Renqiang and Shen, Dinghan and Malon, Christopher and Zhang, Yizhe and Li, Yitong and Carin, Lawrence},
  journal={arXiv preprint arXiv:2006.00693},
  year={2020}
}

@inproceedings{zhang2020content,
  title={Content-collaborative disentanglement representation learning for enhanced recommendation},
  author={Zhang, Yin and Zhu, Ziwei and He, Yun and Caverlee, James},
  booktitle={Proceedings of the 14th ACM Conference on Recommender Systems},
  pages={43--52},
  year={2020}
}

@inproceedings{wang2020disentangled,
  title={Disentangled graph collaborative filtering},
  author={Wang, Xiang and Jin, Hongye and Zhang, An and He, Xiangnan and Xu, Tong and Chua, Tat-Seng},
  booktitle={Proceedings of the 43rd international ACM SIGIR conference on research and development in information retrieval},
  pages={1001--1010},
  year={2020}
}

@inproceedings{tsai2018learning,
  title={Learning Factorized Multimodal Representations},
  author={Tsai, Yao-Hung Hubert and Liang, Paul Pu and Zadeh, Amir and Morency, Louis-Philippe and Salakhutdinov, Ruslan},
  booktitle={International Conference on Learning Representations},
  year={2018}
}

@article{shi2019variational,
  title={Variational mixture-of-experts autoencoders for multi-modal deep generative models},
  author={Shi, Yuge and Paige, Brooks and Torr, Philip and others},
  journal={Advances in neural information processing systems},
  volume={32},
  year={2019}
}

@inproceedings{alaniz2022compositional,
  title={Compositional Mixture Representations for Vision and Text},
  author={Alaniz, Stephan and Federici, Marco and Akata, Zeynep},
  booktitle={Proceedings of the IEEE/CVF Conference on Computer Vision and Pattern Recognition},
  pages={4202--4211},
  year={2022}
}

@inproceedings{lee2021dranet,
  title={Dranet: Disentangling representation and adaptation networks for unsupervised cross-domain adaptation},
  author={Lee, Seunghun and Cho, Sunghyun and Im, Sunghoon},
  booktitle={Proceedings of the IEEE/CVF conference on computer vision and pattern recognition},
  pages={15252--15261},
  year={2021}
}

@inproceedings{li2023revisiting,
  title={Revisiting disentanglement and fusion on modality and context in conversational multimodal emotion recognition},
  author={Li, Bobo and Fei, Hao and Liao, Lizi and Zhao, Yu and Teng, Chong and Chua, Tat-Seng and Ji, Donghong and Li, Fei},
  booktitle={Proceedings of the 31st ACM International Conference on Multimedia},
  pages={5923--5934},
  year={2023}
}

@inproceedings{yang2022disentangled,
  title={Disentangled representation learning for multimodal emotion recognition},
  author={Yang, Dingkang and Huang, Shuai and Kuang, Haopeng and Du, Yangtao and Zhang, Lihua},
  booktitle={Proceedings of the 30th ACM International Conference on Multimedia},
  pages={1642--1651},
  year={2022}
}

@inproceedings{perrett2021trx,
  title={Temporal-relational crosstransformers for few-shot action recognition},
  author={Perrett, Toby and Masullo, Alessandro and Burghardt, Tilo and Mirmehdi, Majid and Damen, Dima},
  booktitle={Proceedings of the IEEE/CVF conference on computer vision and pattern recognition},
  pages={475--484},
  year={2021}
}

@inproceedings{wang2022hyrsm,
  title={Hybrid relation guided set matching for few-shot action recognition},
  author={Wang, Xiang and Zhang, Shiwei and Qing, Zhiwu and Tang, Mingqian and Zuo, Zhengrong and Gao, Changxin and Jin, Rong and Sang, Nong},
  booktitle={Proceedings of the IEEE/CVF Conference on Computer Vision and Pattern Recognition},
  pages={19948--19957},
  year={2022}
}

@article{vaswani2017attention,
  title={Attention is all you need},
  author={Vaswani, Ashish and Shazeer, Noam and Parmar, Niki and Uszkoreit, Jakob and Jones, Llion and Gomez, Aidan N and Kaiser, {\L}ukasz and Polosukhin, Illia},
  journal={Advances in neural information processing systems},
  volume={30},
  year={2017}
}

@book{lucas1981iterative,
  title={An iterative image registration technique with an application to stereo vision},
  author={Lucas, Bruce D and Kanade, Takeo and others},
  volume={81},
  year={1981},
  publisher={Vancouver}
}

@ARTICLE{Shahroudy2018MMaction,
  author={Shahroudy, Amir and Ng, Tian-Tsong and Gong, Yihong and Wang, Gang},
  journal={IEEE Transactions on Pattern Analysis and Machine Intelligence}, 
  title={Deep Multimodal Feature Analysis for Action Recognition in RGB+D Videos}, 
  year={2018},
  volume={40},
  number={5},
  pages={1045-1058},
  doi={10.1109/TPAMI.2017.2691321}}

@InProceedings{Joze2020MMTM,
author = {Joze, Hamid Reza Vaezi and Shaban, Amirreza and Iuzzolino, Michael L. and Koishida, Kazuhito},
title = {MMTM: Multimodal Transfer Module for CNN Fusion},
booktitle = {Proceedings of the IEEE/CVF Conference on Computer Vision and Pattern Recognition (CVPR)},
month = {June},
year = {2020}
}

@inproceedings{gu20203d,
  title={3d hand pose estimation with disentangled cross-modal latent space},
  author={Gu, Jiajun and Wang, Zhiyong and Ouyang, Wanli and Li, Jiafeng and Zhuo, Li and others},
  booktitle={Proceedings of the IEEE/CVF Winter Conference on Applications of Computer Vision},
  pages={391--400},
  year={2020}
}

@inproceedings{yao2024drfuse,
  title={DrFuse: Learning Disentangled Representation for Clinical Multi-Modal Fusion with Missing Modality and Modal Inconsistency},
  author={Yao, Wenfang and Yin, Kejing and Cheung, William K and Liu, Jia and Qin, Jing},
  booktitle={Proceedings of the AAAI Conference on Artificial Intelligence},
  volume={38},
  number={15},
  pages={16416--16424},
  year={2024}
}

@article{paszke2019pytorch,
  title={Pytorch: An imperative style, high-performance deep learning library},
  author={Paszke, Adam and Gross, Sam and Massa, Francisco and Lerer, Adam and Bradbury, James and Chanan, Gregory and Killeen, Trevor and Lin, Zeming and Gimelshein, Natalia and Antiga, Luca and others},
  journal={Advances in neural information processing systems},
  volume={32},
  year={2019}
}

@inproceedings{chen2019temporal,
  title={Temporal attentive alignment for large-scale video domain adaptation},
  author={Chen, Min-Hung and Kira, Zsolt and AlRegib, Ghassan and Yoo, Jaekwon and Chen, Ruxin and Zheng, Jian},
  booktitle={Proceedings of the IEEE/CVF International Conference on Computer Vision},
  pages={6321--6330},
  year={2019}
}

@article{gao2020pairwise,
  title={A pairwise attentive adversarial spatiotemporal network for cross-domain few-shot action recognition-R2},
  author={Gao, Zan and Guo, Leming and Guan, Weili and Liu, An-An and Ren, Tongwei and Chen, Shengyong},
  journal={IEEE Transactions on Image Processing},
  volume={30},
  pages={767--782},
  year={2020},
  publisher={IEEE}
}

@inproceedings{kim2022learning,
  title={Learning mixture of domain-specific experts via disentangled factors for autonomous driving},
  author={Kim, Inhan and Lee, Joonyeong and Kim, Daijin},
  booktitle={Proceedings of the AAAI Conference on Artificial Intelligence},
  volume={36},
  number={1},
  pages={1148--1156},
  year={2022}
}

@article{10.1109/TMM.2023.3321430,
author = {Lu, Yuwu and Wong, Wai Keung and Yuan, Chun and Lai, Zhihui and Li, Xuelong},
title = {Low-Rank Correlation Learning for Unsupervised Domain Adaptation},
year = {2023},
issue_date = {2024},
publisher = {IEEE Press},
volume = {26},
issn = {1520-9210},
url = {https://doi.org/10.1109/TMM.2023.3321430},
doi = {10.1109/TMM.2023.3321430},
journal = {IEEE Transactions on Multimedia (TMM)},
month = oct,
pages = {4153–4167},
numpages = {15}
}

@article{10.1109/TMM.2024.3361729,
author = {Yuan, Jin and Hou, Feng and Yang, Ying and Zhang, Yang and Shi, Zhongchao and Geng, Xin and Fan, Jianping and He, Zhiqiang and Rui, Yong},
title = {Domain-Aware Graph Network for Bridging Multi-Source Domain Adaptation},
year = {2024},
issue_date = {2024},
publisher = {IEEE Press},
volume = {26},
issn = {1520-9210},
journal = {IEEE Transactions on Multimedia (TMM)},
month = feb,
pages = {7210–7224},
numpages = {15}
}

@article{10.1109/TMM.2012.2229263,
author = {Ozkan, Derya and Morency, Louis-Philippe},
title = {Latent Mixture of Discriminative Experts},
year = {2013},
issue_date = {February 2013},
publisher = {IEEE Press},
volume = {15},
number = {2},
issn = {1520-9210},
url = {https://doi.org/10.1109/TMM.2012.2229263},
doi = {10.1109/TMM.2012.2229263},
journal = {IEEE Transactions on Multimedia (TMM)},
month = feb,
pages = {326–338},
numpages = {13}
}

@article{10.1109/TMM.2024.3384058,
author = {Yi, Jing and Chen, Zhenzhong},
title = {Variational Mixture of Stochastic Experts Auto-Encoder for Multi-Modal Recommendation},
year = {2024},
issue_date = {2024},
publisher = {IEEE Press},
volume = {26},
issn = {1520-9210},
url = {https://doi.org/10.1109/TMM.2024.3384058},
doi = {10.1109/TMM.2024.3384058},
journal = {IEEE Transactions on Multimedia (TMM)},
month = apr,
pages = {8941–8954},
numpages = {14}
}

@article{10.1109/TMM.2024.3399468,
author = {Xu, Fangbin and Chen, Dongyue and Jia, Tong and Deng, Shizhuo and Wang, Hao},
title = {CBDMoE: Consistent-but-Diverse Mixture of Experts for Domain Generalization},
year = {2024},
issue_date = {2024},
publisher = {IEEE Press},
volume = {26},
issn = {1520-9210},
url = {https://doi.org/10.1109/TMM.2024.3399468},
doi = {10.1109/TMM.2024.3399468},
journal = {IEEE Transactions on Multimedia (TMM)},
month = may,
pages = {9814–9824},
numpages = {11}
}

@article{10.1109/TMM.2024.3360710,
author = {Zhou, Xiaofei and Wu, Zhicong and Cong, Runmin},
title = {Decoupling and Integration Network for Camouflaged Object Detection},
year = {2024},
issue_date = {2024},
publisher = {IEEE Press},
volume = {26},
issn = {1520-9210},
url = {https://doi.org/10.1109/TMM.2024.3360710},
doi = {10.1109/TMM.2024.3360710},
journal = {IEEE Transactions on Multimedia (TMM)},
month = jan,
pages = {7114–7129},
numpages = {16}
}

@article{10.1109/TMM.2023.3347645,
author = {Liu, Yabo and Wang, Jinghua and Wang, Weijia and Hu, Yu and Wang, Yaowei and Xu, Yong},
title = {CRADA: Cross Domain Object Detection With Cyclic Reconstruction and Decoupling Adaptation},
year = {2024},
issue_date = {2024},
publisher = {IEEE Press},
volume = {26},
issn = {1520-9210},
url = {https://doi.org/10.1109/TMM.2023.3347645},
doi = {10.1109/TMM.2023.3347645},
journal = {IEEE Transactions on Multimedia (TMM)},
month = jan,
pages = {6250–6261},
numpages = {12}
}

@article{10.1109/TMM.2023.3265843,
author = {Li, Hua and Liang, Junyan and Wu, Ruiqi and Cong, Runmin and Wu, Wenhui and Wu Kwong, Sam Tak},
title = {Stereo Superpixel Segmentation via Decoupled Dynamic Spatial-Embedding Fusion Network},
year = {2023},
issue_date = {2024},
publisher = {IEEE Press},
volume = {26},
issn = {1520-9210},
url = {https://doi.org/10.1109/TMM.2023.3265843},
doi = {10.1109/TMM.2023.3265843},
journal = {IEEE Transactions on Multimedia (TMM)},
month = apr,
pages = {367–378},
numpages = {12}
}

@article{markham2024understanding,
  title={Understanding the Cross-Domain Capabilities of Video-Based Few-Shot Action Recognition Models},
  author={Markham, Georgia and Balamurali, Mehala and Hill, Andrew J},
  journal={arXiv preprint arXiv:2406.01073},
  year={2024}
}

@article{wang2024tamt,
  title={TAMT: Temporal-Aware Model Tuning for Cross-Domain Few-Shot Action Recognition},
  author={Wang, Yilong and Gao, Zilin and Wang, Qilong and Chen, Zhaofeng and Li, Peihua and Hu, Qinghua},
  journal={arXiv preprint arXiv:2411.19041},
  year={2024}
}

@article{guo2025dmsd,
  title={DMSD-CDFSAR: Distillation from mixed-source domain for cross-domain few-shot action recognition},
  author={Guo, Fei and Wang, YiKang and Qi, Han and Zhu, Li and Sun, Jing},
  journal={Expert Systems with Applications},
  pages={126411},
  year={2025},
  publisher={Elsevier}
}

@inproceedings{qu2024lead,
  title={Lead: Learning decomposition for source-free universal domain adaptation},
  author={Qu, Sanqing and Zou, Tianpei and He, Lianghua and R{\"o}hrbein, Florian and Knoll, Alois and Chen, Guang and Jiang, Changjun},
  booktitle={Proceedings of the IEEE/CVF Conference on Computer Vision and Pattern Recognition},
  pages={23334--23343},
  year={2024}
}

\begin{IEEEbiography}
[{\includegraphics[width=1in,height=1.25in,clip,keepaspectratio]{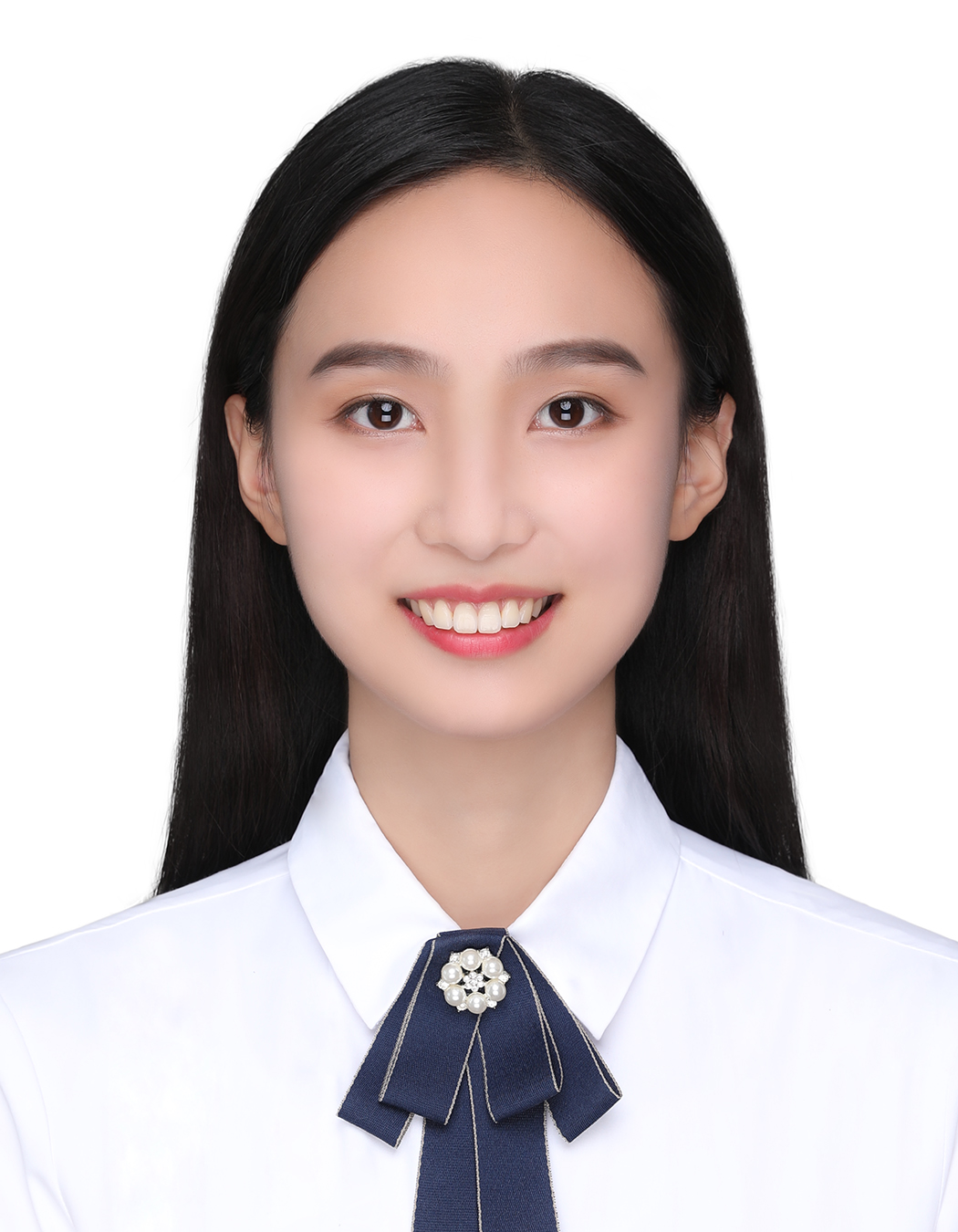}}]{Yuyang Wanyan} is currently pursuing the Ph.D. degree with the Multimedia Computing Group, State Key Laboratory of Multimodal Artificial Intelligence Systems (MAIS), Institute of Automation, Chinese Academy of Sciences, and the School of Artificial Intelligence at the University of Chinese Academy of Sciences. She received her B.Sc. degree in Computer Science and Technology from Jilin University in 2021. Her research interests include multimedia analysis, computer vision, and the multimodal large language model. 
\end{IEEEbiography}

\begin{IEEEbiography}
[{\includegraphics[width=1in,height=1.25in,clip,keepaspectratio]{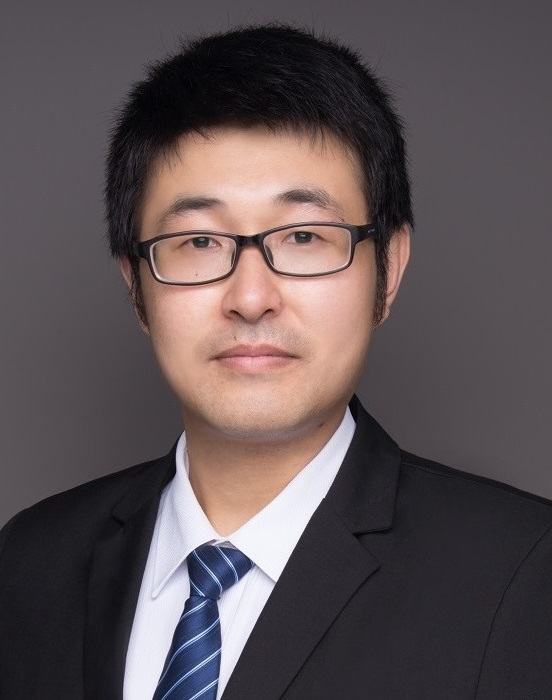}}]{Xiaoshan Yang}
received the Ph.D. degree from the Institute of Automation, Chinese Academy of Sciences, Beijing, China, in 2016. He is currently a Professor with the Institute of Automation, Chinese Academy of Sciences. His research interests include multimedia analysis and computer vision. He won the President Award of the Chinese Academy of Sciences in 2016, the Excellent Doctoral Dissertation of the Chinese Academy of Sciences in 2017, and the CCF-Tencent Rhino Bird Excellence Award in 2018. He was an area chair of ACM MM/IJCAI/ICPR, and reviewer for several top-tier journals and conferences, e.g., TPAMI, PR, IJCV, ACM MM, CVPR, ICCV.
\end{IEEEbiography}

\begin{IEEEbiography}
[{\includegraphics[width=1in,height=1.25in,clip,keepaspectratio]{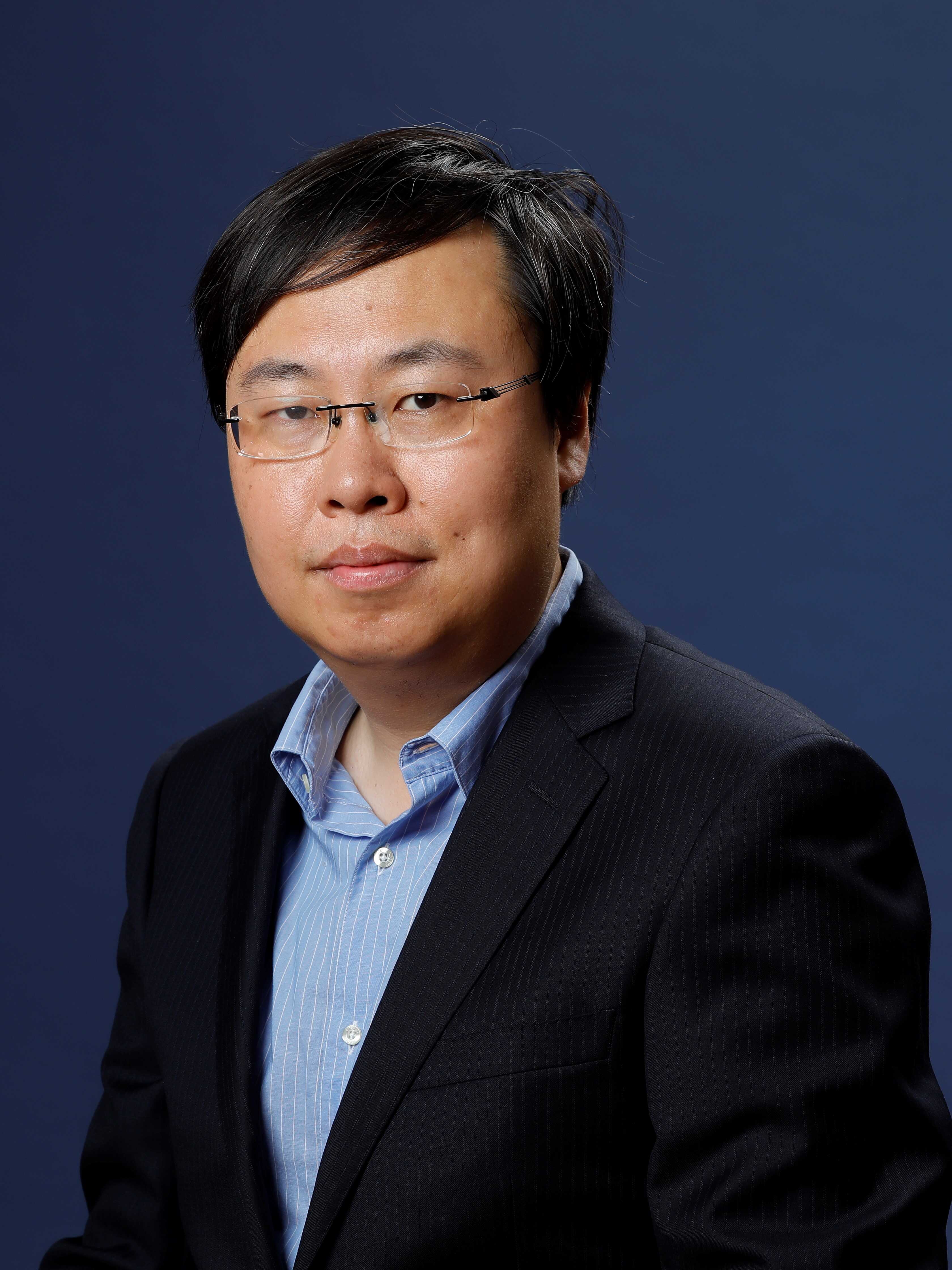}}]{Weiming Dong} (Member, IEEE) is a Professor at the State Key Laboratory of Multimodal Artificial Intelligence Systems (MAIS), Institute of Automation, Chinese Academy of Sciences.   He received his BSc and MSc degrees in 2001 and 2004, both from Tsinghua University, China.   He received his PhD in Computer Science from the University of Lorraine, France, in 2007. His research interests include image synthesis, image recognition, and computational creativity.
\end{IEEEbiography}

\begin{IEEEbiography}
[{\includegraphics[width=1in,height=1.25in,clip,keepaspectratio]{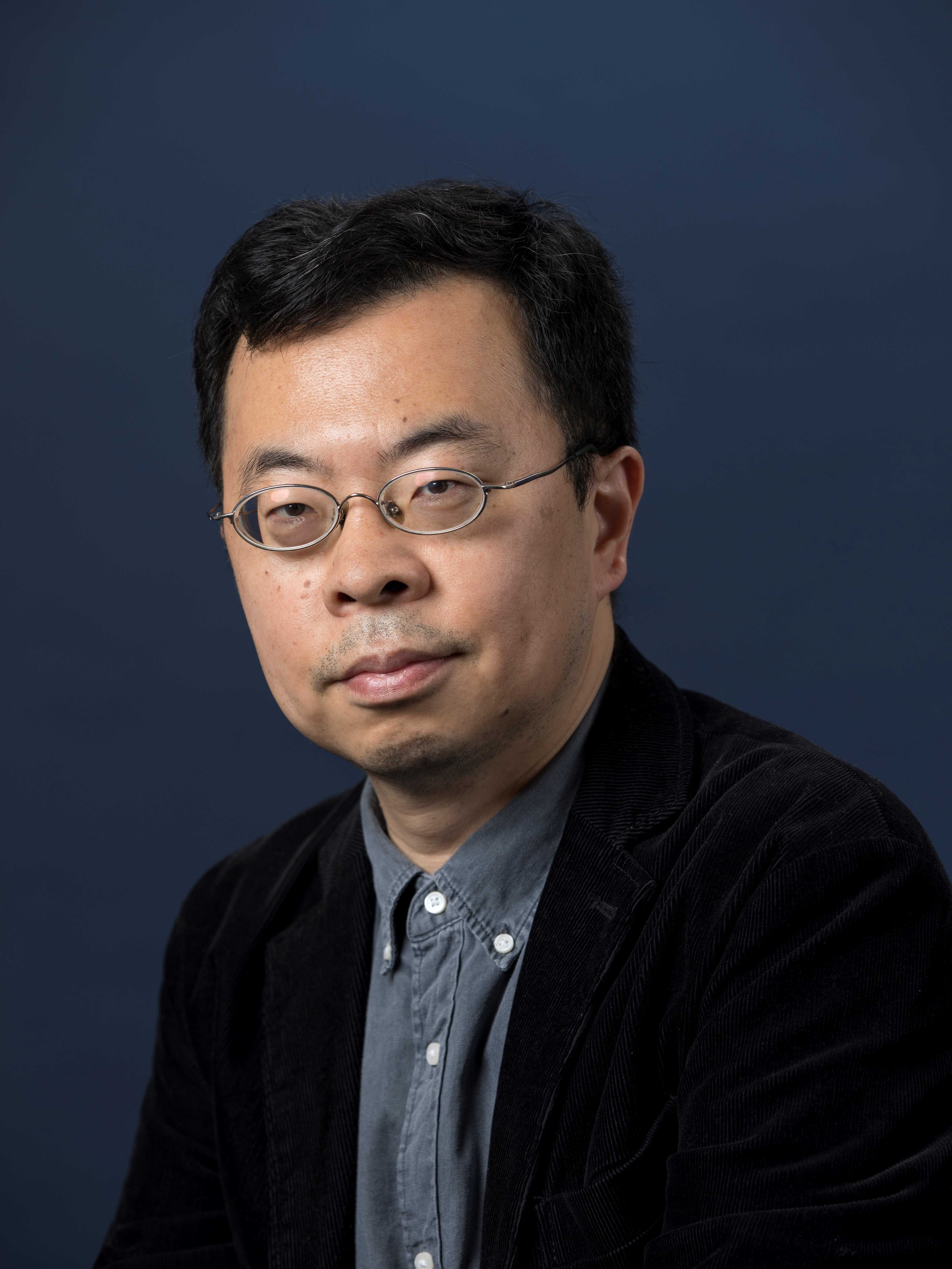}}]{Changsheng Xu}
(Fellow, IEEE) is a Professor at the State Key Laboratory of Multimodal Artificial Intelligence Systems (MAIS), Institute of Automation, Chinese Academy of Sciences. His research interests include multimedia content analysis/indexing/retrieval, pattern recognition and computer vision. He has held 50 granted/pending patents and published over 400 refereed research papers in these areas. Dr. Xu has served as associate editor, guest editor, general chair, program chair, area/track chair and TPC member for over 20 IEEE and ACM prestigious multimedia journals, conferences and workshops, including IEEE Trans. on Multimedia, ACM Trans. on Multimedia Computing, Communications and Applications and ACM Multimedia conference. He is IEEE Fellow, IAPR Fellow and ACM Distinguished Scientist. 
\end{IEEEbiography}

\end{document}